\pdfoutput=1
\documentclass[sigconf]{acmart}
\acmConference[]{}{}{}
\renewcommand\footnotetextcopyrightpermission[1]{}
\usepackage[normalem]{ulem}

\usepackage{microtype}

\usepackage[utf8]{inputenc}
\usepackage{amsmath}
\usepackage{amsthm}
\usepackage{xspace}
\usepackage{varioref}
\usepackage{graphicx}
\usepackage{subcaption}
\usepackage{amsfonts}
\usepackage{booktabs}
\usepackage{multirow}
\usepackage{etoolbox}
\usepackage{enumitem}
\pretocmd{\eqref}{Eq.~}{}{}
\usepackage{mathtools}
\usepackage[frozencache,cachedir=.]{minted} 
\usepackage{algorithm}
\usepackage{algorithmic}
\usepackage{esvect}
\usepackage{booktabs}
\usepackage[capitalize]{cleveref}
\setminted{fontfamily=zlmtt}
\DeclarePairedDelimiter\ceil{\lceil}{\rceil}
\DeclarePairedDelimiter\floor{\lfloor}{\rfloor}
\newcommand{\presec}{}
\newcommand{\postsec}{}
\newcommand{\presub}{}

\newcommand{\sys}{\textsc{ALT}\xspace}
\begin{document}
\title{\sys: Boosting Deep Learning Performance by Breaking the Wall between Graph and Operator Level Optimizations}
\author{Zhiying Xu}
\email{zyxu@smail.nju.edu.cn}
\affiliation{%
	\institution{Nanjing University}
	\country{China}
}
\author{Jiafan Xu}
\email{mf20330099@smail.nju.edu.cn}
\affiliation{%
	\institution{Nanjing University}
	\country{China}
}
\author{Hongding Peng}
\email{mg20330044@smail.nju.edu.cn}
\affiliation{%
	\institution{Nanjing University}
	\country{China}
}
\author{Wei Wang}
\email{ww@nju.edu.cn}
\affiliation{%
	\institution{Nanjing University}
	\country{China}
}
\author{Xiaoliang Wang}
\email{waxili@nju.edu.cn}
\affiliation{%
	\institution{Nanjing University}
	\country{China}
}
\author{Haoran Wan}
\email{wanhr@smail.nju.edu.cn}
\affiliation{%
	\institution{Nanjing University}
	\country{China}
}
\author{Haipeng Dai}
\email{haipengdai@nju.edu.cn}
\affiliation{%
	\institution{Nanjing University}
	\country{China}
}
\author{Yixu Xu}
\email{xuyixu@huawei.com}
\affiliation{
	\institution{Huawei Technologies}
	\country{China}
}
\author{Hao Cheng}
\email{chenghao49@hisilicon.com}
\affiliation{
	\institution{Huawei Technologies}
	\country{China}
}
\author{Kun Wang}
\email{kun.wang1981@gmail.com}
\affiliation{%
	\institution{The Hong Kong Polytechnic University}
	\country{China}
}
\author{Guihai Chen}
\email{gchen@nju.edu.cn}
\affiliation{%
	\institution{Nanjing University}
	\country{China}
}

\begin{abstract}
%
Deep learning models rely on highly optimized tensor libraries for efficient inference on heterogeneous hardware.
%
%
Current deep compilers typically predetermine layouts of tensors and then optimize loops of operators. However, such unidirectional and one-off workflow strictly separates graph-level optimization and operator-level optimization into different system layers, missing opportunities for unified tuning.
%

This paper proposes \sys, a compiler that performs joint graph- and operator-level optimizations for deep models. 
\sys provides a generic transformation module to manipulate layouts and loops with easy-to-use primitive functions.
%
\sys further integrates an auto-tuning module that jointly optimizes graph-level data layouts and operator-level loops while guaranteeing efficiency. 
%
%
%
Experimental results show that \sys significantly outperforms state-of-the-art compilers (\textit{e.g.}, Ansor) in terms of both single operator performance (\textit{e.g.}, $1.5\times$ speedup on average) and end-to-end inference performance (\textit{e.g.}, $1.4\times$ speedup on average).
\end{abstract}

\maketitle
\thispagestyle{empty}
 \section{Introduction} 
\label{sec:intro}
Deep learning has become one of the essential building blocks for emerging applications, such as machine translation and autonomous driving systems. To provide ubiquitous services, 
%
developers craft high-performance programs supporting various tensor operators (\textit{e.g.}, 2-D convolution and matrix multiplication) on different hardware platforms (\textit{e.g.}, NVIDIA GPU and ARM CPU).
However, current vendor libraries (\textit{e.g.}, MKL-DNN \cite{mkldnn} and cuDNN \cite{chetlur2014cudnn}) typically demand significant engineering effort on manual optimization.
%
Moreover, the hand-tuning approach can hardly catch up with the fast evolution of deep learning techniques that constantly introduce new tensor operators \cite{ho2022video} and new hardware (\textit{e.g.}, neural processing units).
%
Therefore, researchers develop deep compilers \cite{leary2017xla, chen2018tvm, vasilache2018tensor, baghdadi2019tiramisu, zhao2021akg} to achieve automatic performance optimization by auto-tuning and code generation techniques. 
%

Two key categories of optimizations during compilation are graph-level optimization and operator-level optimization. 
Graph-level optimization represents operators as nodes and tensors as edges in a computational graph and rewrites nodes or edges to obtain a more efficient graph for inference. 
For instance, data layout optimization rewrites the tensor storage format to improve memory accessing performance \cite{ju1991reduction, bacon1994compiler, raman2007structure, cho2008compiler, vasilache2012joint, sharma2015data, shirako2019integrating}. 
Constant folding \cite{muchnick1997advanced, roesch2018relay, boemer2019ngraph} and common subexpression elimination \cite{muchnick1997advanced, roesch2018relay} removes redundant nodes.
Operator-level optimization, which mainly involves loop optimization, transforms the nested loops in the source code of each operator to schedule the execution of instructions \cite{banerjee2007loop, hall2009loop, ragan2013halide, gong2018empirical, chen2018tvm}. 
In this work, we focus on data layout optimization and loop optimization because they yield significant performance improvements and their tuning strongly correlates with operator and hardware characteristics.
%
%
%
%

Unfortunately, existing deep compilers (\textit{e.g.}, TVM \cite{chen2018tvm}, Tensor Comprehension \cite{vasilache2018tensor}, Tiramisu \cite{baghdadi2019tiramisu}, AKG \cite{zhao2021akg}) and auto-tuning techniques (\textit{e.g.}, AutoTVM \cite{chen2018learning}, NeoCPU \cite{liu2019optimizing}, FlexTensor \cite{zheng2020flextensor} and Ansor \cite{zheng2020ansor}), fail to combine data layout and loop optimizations effectively. 
These systems first predetermine tensor layouts either manually or via setting a hyper-parameter from a predefined template and then perform loop optimization based on these layouts.
There are three major limitations in this unidirectional and one-off workflow. First, manual layout selection implies that only a limited number of layout choices can be explored, hence prone to be suboptimal. Second, altering the tensor layout demands the time-consuming re-implementation of operators that access the tensor. Third, layout optimization and loop optimization are separated into different system layers. Such strict boundary seriously compromises the performance of the generated tensor programs. For instance, we observe that optimizing loops based on the best of three candidate layouts for 2-D convolutional operators can improve the performance by $55.9\%$ on average on the Intel CPU. Moreover, the performance of a specific data layout is sensitive to operator configurations (\textit{e.g.}, tensor shapes) and hardware. Thus it is hard to determine data layouts for each workload without feedback from loop optimization. 

This paper proposes \sys, a deep compiler that jointly performs graph-level and operator-level optimizations for deep models. 
The design of \sys originates from the following insight. Graph-level data layout optimization and operator-level loop optimization could benefit from each other. In the meanwhile, the root cause of the inability to perform cross-layer joint tuning is the coupling between data storage and operator implementation in prior arts, such that altering the data layout requires re-implementing operators. Such high cost for changing layouts further leads to the unidirectional and one-off optimization flow. Therefore, \sys abstracts layout manipulation as easy-to-use primitive functions, such that the task of re-implementing operators can be delegated to a compilation pass without human interference. After reducing the cost, \sys further incorporates layout and loop optimizations into a unified auto-tuning framework, which breaks the boundary between graph- and operator-level optimizations to open new opportunities. 

%
%
%
%

It is not trivial to achieve our goals. We need to strike the following challenges.

\noindent\textit{Challenge 1: How can we eliminate the overhead of layout transformation?} We discover two types of potential overhead when altering the tensor layouts: \emph{layout-conversion} overhead and \emph{fusion-conflict} overhead. 
First, operators along the data stream may require different tensor layouts to reach their optimal performance. 
%
%
To transform the layouts of tensors produced by other operators at runtime, 
%
directly inserting conversion operators will incur extra overhead on data movements. 
Second, altering the output tensor layout of an operator needs the reconstruction of its loop nest. 
Such reconstruction may inhibit the operator from being fused with its consumer, which is an important loop optimization technique to improve inter-operator data locality.
%

\noindent\textit{Challenge 2: How can we prevent inefficiency due to the search space reconstruction during joint tuning?} Changing the output layout of an operator will induce the loop nest reconstruction, which will further lead to the variation of the loop tuning space. 
For joint tuning, such space variation prohibits a direct iterative exploration. Otherwise, the points we have searched in the last iteration may be invalid in the changing space. 
%
This leads to inefficient tuning for most search methods, including genetic and learning-based algorithms, since the accumulated knowledge of the search space structure cannot be further exploited in the newly reconstructed space. 

\noindent\textit{Challenge 3: How can we improve efficiency given the search space explosion with the combination of layout and loop tuning?} Along with the joint tuning, the combined search space will be tremendously large, hence inefficient to explore directly. For example, in a typical 2-D convolutional operator, the loop transformation space can contain up to ${O}(10^7)$ points for its seven nested loops. After combining the layout transformation, the joint search space can be at a scale of ${O}(10^{19})$ considering three tensors, each of which further involves four dimensions. Moreover, end-to-end optimization is more challenging due to the inter-dependency of many operators and tensors.

To eliminate the two types of overhead brought by layout transformation, we propose a \emph{layout propagation} mechanism. 
Suppose we have chosen a different layout for the input tensor of an operator.
We let the upstream operator, which is the producer of this tensor, directly yield elements based on this new layout, hence no conversion operator is required.
To promote operator fusion, we propagate the new layout downstream along the computational graph to let the consumer operator trigger the same loop reconstruction, which helps to align loop nests of multiple operators for fusion. As such, we can safely transform data layouts with minimal overhead, and without sabotaging loop optimization. 

To alleviate the search space reconstruction issue in the co-tuning, our solution is two folds. First, we divide the co-tuning into two stages: \emph{joint} stage and \emph{loop-only} stage. The joint stage searches for optimal tensor layouts, while the loop-only stage only performs loop tuning with the searched layouts remaining unchanged. Second, we propose a \emph{cross-exploration architecture} for the joint stage, rather than the direct exploration. For a new feasible layout, we reconstruct the loop space and then perform multiple rounds of loop optimization to assess the new layout. This design avoids inefficient loop space reconstruction since the loop-only stage keeps layouts unchanged. It also achieves the expected bidirectional and unified tuning flow in the joint stage, because each candidate layout is evaluated based on feedback from loop optimization through our novel tuning architecture.

To avoid the search space explosion due to the combination of layout and loop tuning, we prune the space at two levels. First, we only build layout transformation spaces for tensors accessed by complex operators. In this work, we take convolutions and general matrix multiplication as complex operators, the performance of which are layout sensitive. For other tensors, we further exploit the layout propagation mechanism to propagate the searched layouts onto them without more searching. Second, we identify a promising subspace by tailoring a tuning template for each tensor accessed by complex operators. These templates are constructed based on our analysis of layout optimization considering both operator and hardware characteristics.

By addressing these challenges, \sys achieves joint and efficient graph-level data layout optimization and operator-level loop optimization automatically. 

We comprehensively evaluate \sys on Intel CPU, NVIDIA GPU, and ARM CPU. Compared with state-of-the-art vendor libraries (\textit{e.g.}, MKL-DNN \cite{mkldnn}, cuDNN \cite{chetlur2014cudnn}, and XNNPACK \cite{xnnpack}) and auto-tuning frameworks (\textit{e.g.}, Ansor \cite{zheng2020ansor}), \sys achieves an average of $1.5\times$ speedup in terms of single operator performance, and $1.4\times$ speedup in terms of end-to-end inference performance. Our evaluation also shows that \sys can find data layouts that are not explored in prior arts. Additionally, we have deployed \sys in production environments for four months, boosting a broad spectrum of real workloads (\textit{e.g.}, speech recognition and super resolution).

\begin{figure*}[htbp]
\begin{subfigure}[b]{0.3\textwidth}\includegraphics[width=\textwidth]{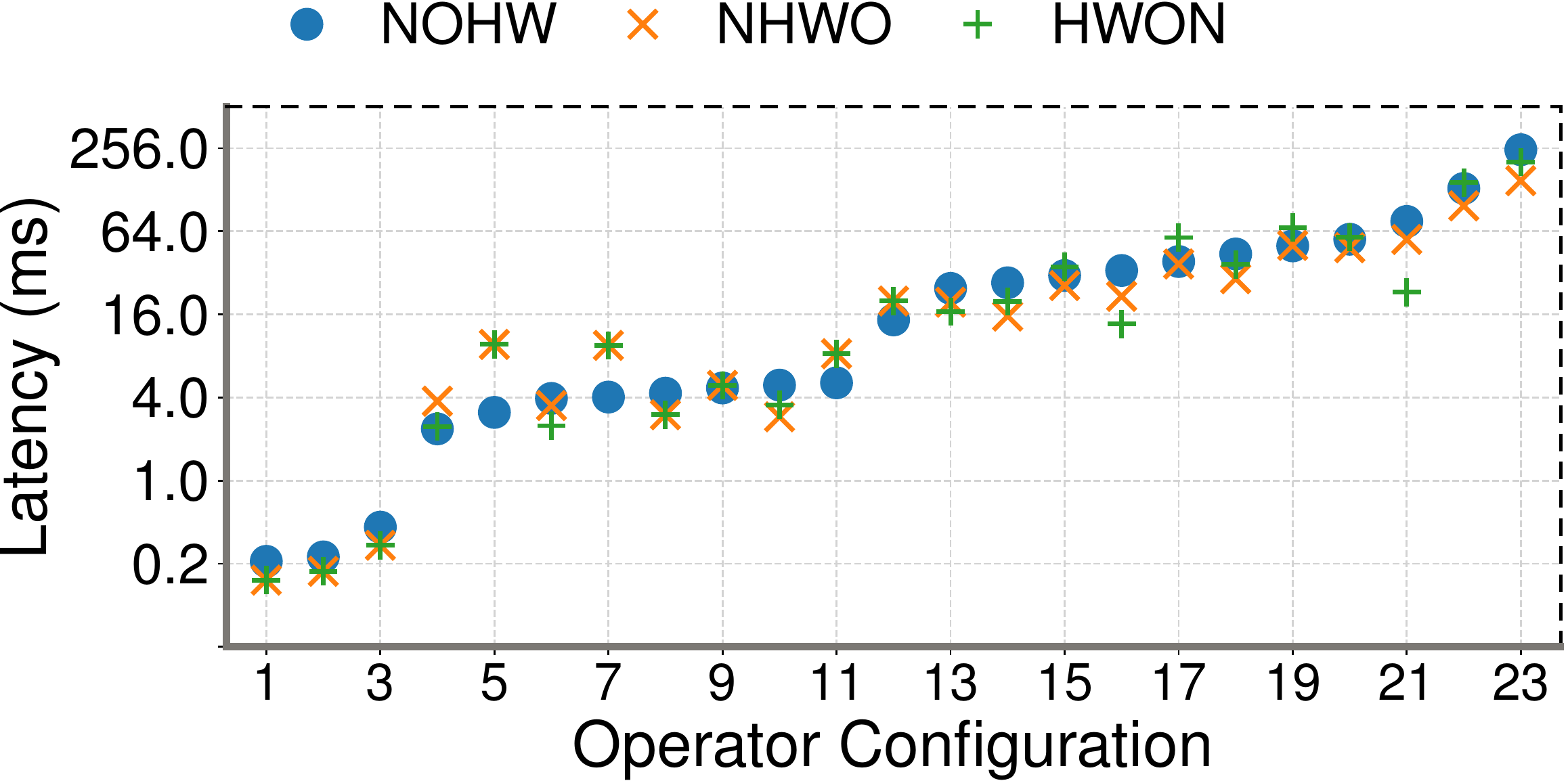}
\caption{C2D on Intel CPU.}
\label{fig:motiv_C2D_llvm}
\end{subfigure}
~
\begin{subfigure}[b]{0.3\textwidth}
\includegraphics[width=\textwidth]{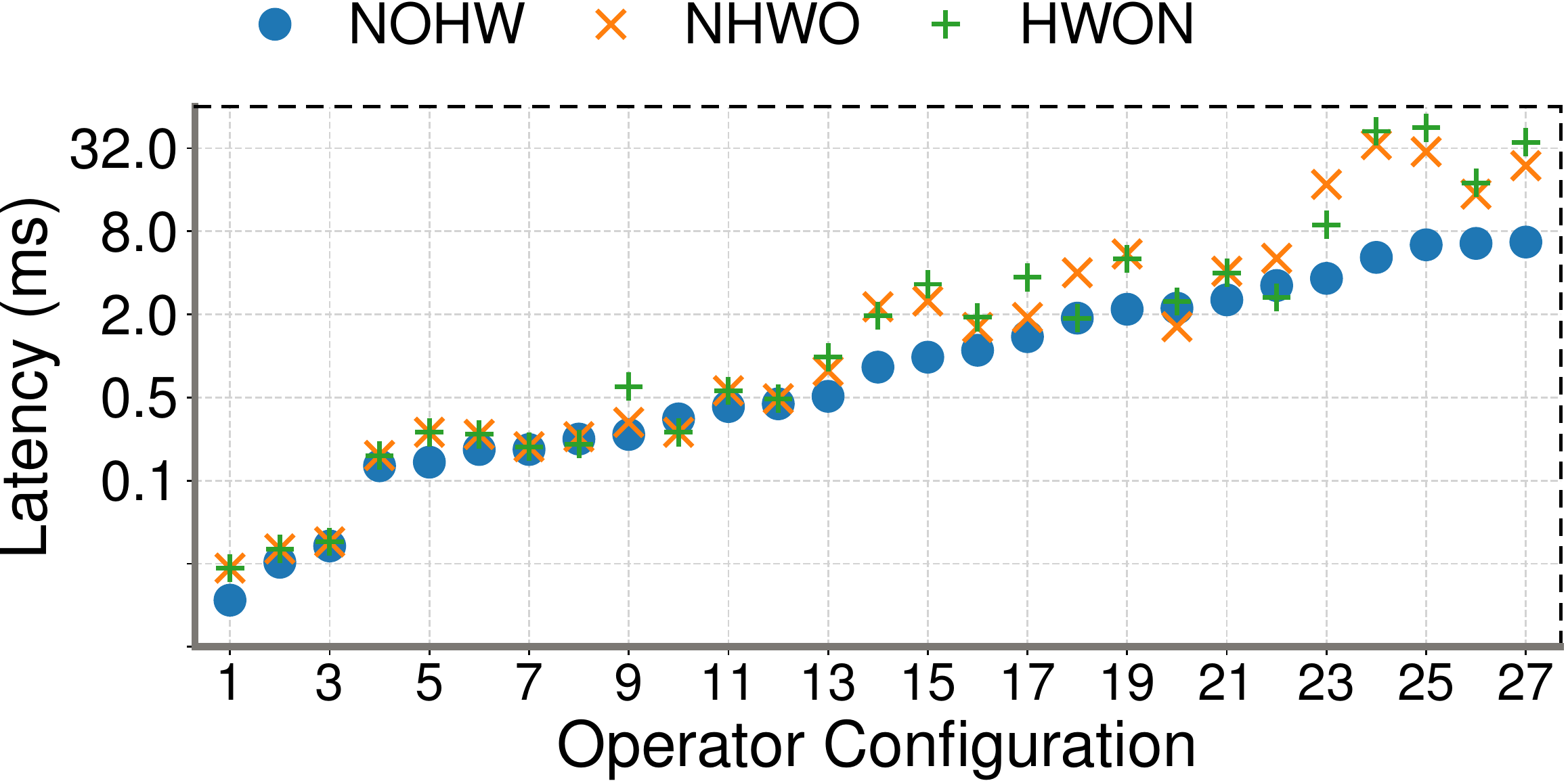}
\caption{C2D on NVIDIA GPU.}
\label{fig:motiv_C2D_cuda}

\end{subfigure}
~
\begin{subfigure}[b]{0.3\textwidth}
\includegraphics[width=\textwidth]{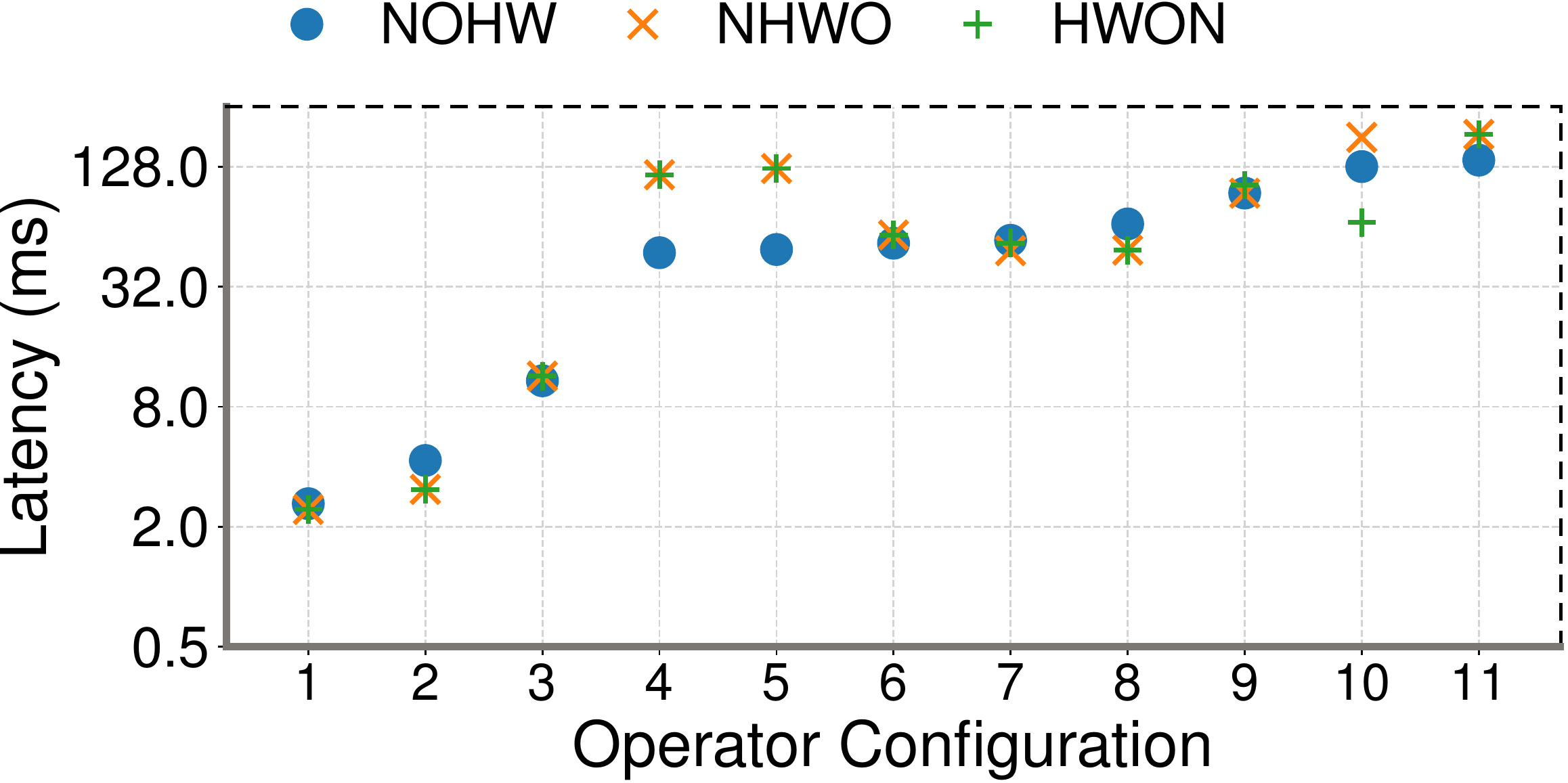}
\caption{C2D on ARM CPU.}
\label{fig:motiv_C2D_android}
\end{subfigure}
\vspace{-0.05in}
\caption{C2D latency with different data layouts on different hardware platforms.}
\label{fig:motiv_C2D}
\vspace{-0.1in}
\end{figure*}

In summary, we make the following contributions:

\begin{itemize}[leftmargin=*]
\item We reveal the necessity of joint graph- and operator-level optimizations for deep learning compilation, and that the root cause of the inefficient unidirectional and one-off optimization flow in prior arts lies in the high cost of layout manipulation.

\item  We design an easy-to-use generic infrastructure that covers a rich layout transformation space. It allows users to manipulate layouts without soiling the hands for re-implementation, and without extra overhead via the layout propagation mechanism during end-to-end optimization.

\item  We devise a joint layout and loop auto-tuning framework. Via effective space pruning and judicious exploration design, it not only achieves a bidirectional and unified optimization flow but also guarantees tuning efficiency.
    
\item  Our extensive evaluation shows that, without human interference, \sys improves performance over state-of-the-art baselines significantly, which also verifies the effectiveness of the proposed techniques.
\end{itemize}
\presec \section{Background and Motivation}
\label{sec:motiv}
A deep compiler typically compiles a neural network with multi-stage lowering and optimization. The compiler takes a model that can be generated by other frameworks (\textit{e.g.}, TensorFlow \cite{abadi2016tensorflow}) as input. It then resolves the model to a computational graph where operators and tensors are represented as nodes and edges, respectively.

Data layout optimization \cite{ju1991reduction, bacon1994compiler, raman2007structure, cho2008compiler, vasilache2012joint, sharma2015data, shirako2019integrating} is to rewrite the tensor storage format (\textit{i.e.}, the attributes of an edge) to alleviate memory accessing overhead for operators that access the tensor. Thus, data layout optimization is often classified as graph-level optimization. The storage format refers to the arrangement of tensor dimensions. Take the 2-D convolution (C2D) operator as an example. Popular data layouts for the output tensor of C2D include $NOHW$, $NHWO$, and $HWON$, where $N, O, H, W$ represent the batch size, the number of output channels, the output tensor height, and the output tensor width, respectively. $NOHW$ is widely used on GPU \cite{paszke2019pytorch}, $NHWO$ is the default layout on CPU in TensorFlow \cite{abadi2016tensorflow}, and $HWON$ is used in digital signal processing. 

After graph-level optimization, the compiler will lower each node in the computational graph to operator-level representation. An operator can typically be represented as deeply nested loops. As the major part of operator-level optimization, loop optimization (\textit{e.g.}, loop tiling, vectorization, etc.) \cite{banerjee2007loop, hall2009loop, ragan2013halide, gong2018empirical, chen2018tvm} is to transform the loop nest to schedule the execution of statements of each operator.

The motivation for this work is as follows.

\noindent\textbf{Observation 1: It is beneficial to jointly perform data layout optimization and loop optimization.} We illustrate the benefits by an experiment that optimizes loops of C2D based on $NOHW$, $NHWO$, and $HWON$ layouts, respectively. Our platforms include 32-core Intel Xeon Silver 4110 CPU@2.1GHz, NVIDIA RTX 2080Ti GPU, and Kirin 990 ARM SoC. We report the performance in \cref{fig:motiv_C2D}, where the latency axis is in log scale, and each hardware involves multiple operator configurations (different number of channels, convolutional strides, etc.) to cover rich workloads. 
We observe that the best layout could improve the performance of loop optimization by 55.9\%, 87.2\%, and 48.8\% on average on Intel CPU, NVIDIA GPU, and ARM CPU, respectively. 
On the converse, making a choice among different layouts is not easy when there is no feedback from loop optimization, due to the highly divergent performance with regard to operator configurations and platforms. For example, although $NHWO$ often outperforms $NOHW$ and $HWON$ on CPUs, especially when the number of input channels is small, there is still no clear rule that can fit all configurations.
%

\noindent\textbf{Observation 2: Existing solutions cannot effectively perform joint tuning due to the high cost of layout manipulation.} Existing systems \cite{chen2018tvm, vasilache2018tensor, baghdadi2019tiramisu} typically couple the tensor storage with the implementation of operators, thus changing layouts requires re-implementation. Such a high cost of layout manipulation limits the number of layout choices that can be explored, and further leads to the unidirectional optimization flow. 
While there are works using special layouts to improve versatility, \emph{e.g.}, $N\frac{O}{o_t}HWo_t$ where $o_t$ is a tiling parameter that can be changed without re-implementation \cite{liu2019optimizing}, they still only cover a small layout optimization space. Moreover, switching to another category of layouts still requires re-implementing operators and even rewriting loop-tuning templates. 
%

\begin{figure}
\centering
\includegraphics[width=0.48\textwidth]{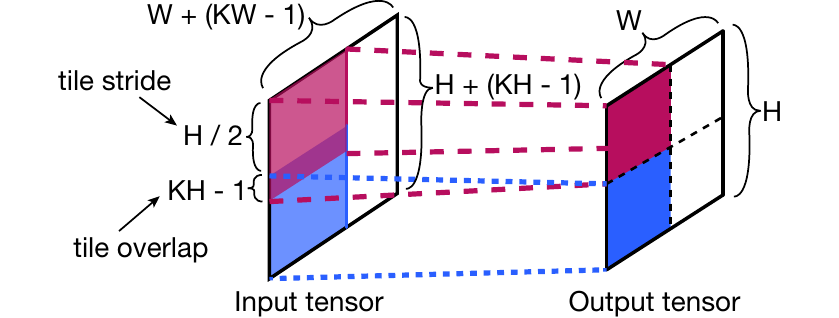}
	\vspace{-0.1in}
	\caption{Layout with overlapped tiling.}
	\label{fig:motiv_layout}
	\vspace{-0.15in}
\end{figure}

\begin{figure}
\centering
\begin{minted}[fontsize=\small]{python}
for n in range(N):
  for oh, ow in range(2, 2):
    for oo in range(O // o_t):
      for ih, iw in range(H // 2, W // 2):
        for io in range(o_t):
          Conv[n][oh][ow][oo][ih][iw][io] = 0.0
        for i, rh, rw in range(I, KH, KW):
          for io in range(o_t):
            Conv[n][oh][ow][oo][ih][iw][io] += \
                Inp[n][oh][ow][i][ih+rh][iw+rw]\
                * Ker[oo][i][rh][rw][io]
\end{minted}
\vspace{-0.09in}
\caption{Program based on the layout in \cref{fig:motiv_layout}.}
\vspace{-0.08in}
\label{fig:motiv_program}

\end{figure}

We use a more versatile layout as a motivating example. This layout is outside the tuning space of $N\frac{O}{o_t}HWo_t$ and is hard to be discovered manually or without joint tuning. It can achieve performance improvement of 32.4\% over $N\frac{O}{o_t}HWo_t$.
%
Besides tiling the channel dimension, this layout further tiles the spatial dimensions (the height and the width) of the output tensor into four blocks.
Each spatial tile of the output tensor has shape $\frac{H}{2}\times \frac{W}{2}$. For a C2D with convolutional stride $1$, the height and the width of the input tensor are $H + (KH - 1)$ and $W + (KW - 1)$, where $KH$ and $KW$ are the height and the width of the convolutional window. Due to the sliding-window operation of C2D that has natural overlaps, each output tile requires a $\left(\frac{H}{2} + (KH - 1)\right)\times \left(\frac{W}{2} + (KW - 1)\right)$ tile of the input tensor for convolution. This leads to the layout in \cref{fig:motiv_layout}, where each colored area denotes a tile, and the overlap between tiles along the input tensor height is exactly $(KH - 1)$.
%
After the layout transformation, the generated loop nest is shown in \cref{fig:motiv_program}, where $I$ is the number of input channels, $Conv$, $Inp$, and $Ker$ are the output tensor, input tensor, and weight tensor, respectively. In \cref{fig:motiv_program}, we also tile the output channels by $o_t$ to achieve multi-dimensional layout tiling. Besides, the corresponding loop $io$ is placed as the innermost loop to improve locality, as a showcase for joint layout and loop optimization. The shape of $Conv$ in \cref{fig:motiv_program} is $N\times2\times2\times \frac{O}{o_t}\times\frac{H}{2}\times\frac{W}{2}\times o_t$.
%
%
Such multi-dimensional tiling with overlaps promotes data locality and cache utilization.
%
We defer the detailed profiling results on various layouts in \cref{subsubsec:case_study}.

%
%
%


\presec\section{System Overview}\postsec
\begin{figure}
	\centering
	\includegraphics[width=0.44\textwidth]{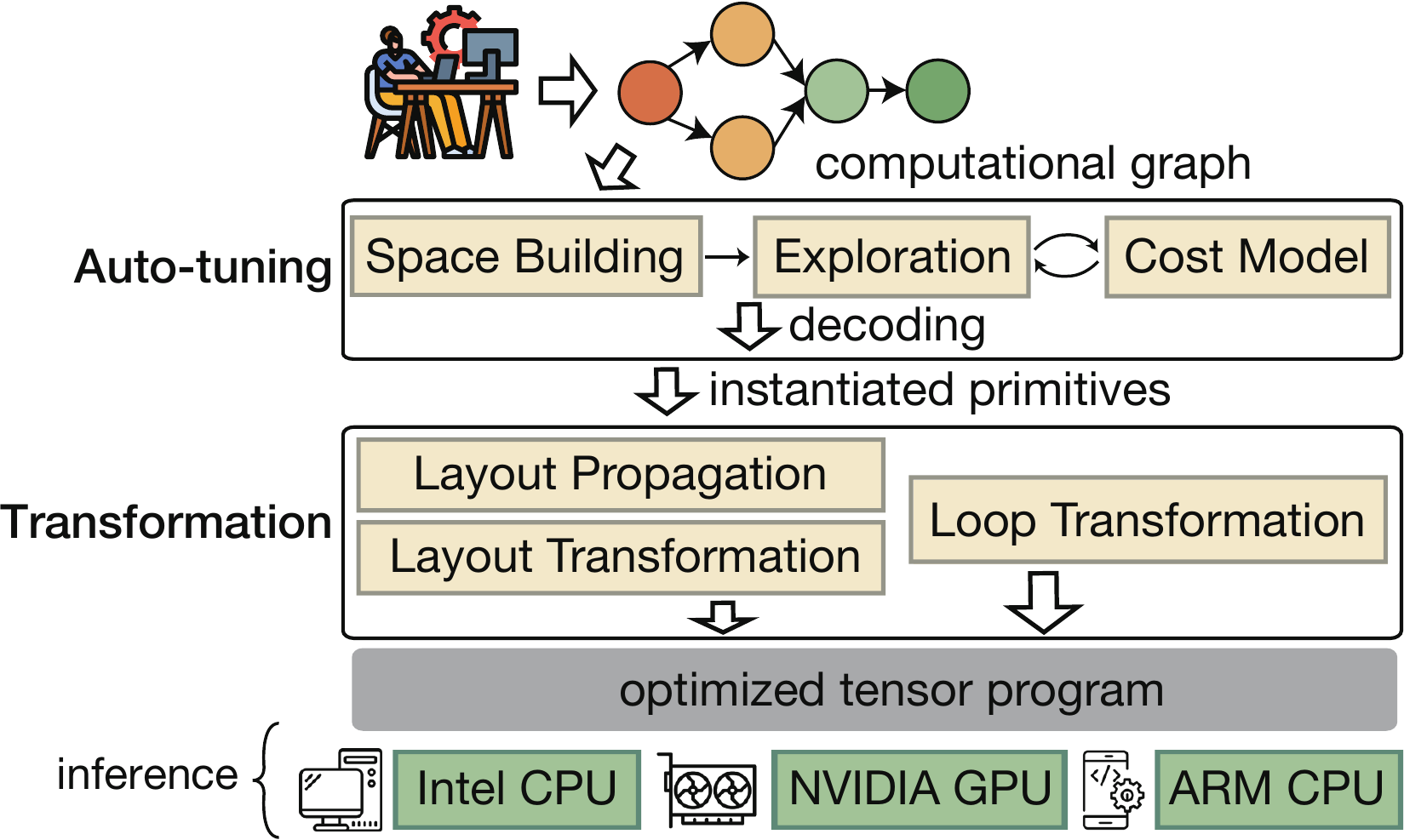}\hfill
	\vspace{-0.05in}
	\caption{Design overview of \sys.}
	\label{fig:design_overview}
	\vspace{-0.1in}
\end{figure}
\sys is a deep compiler that achieves joint graph-level layout optimization and operator-level loop optimization to generate high-performance tensor programs for heterogeneous platforms automatically.
The system overview of \sys is depicted in \cref{fig:design_overview}, which incorporates two major modules: \emph{auto-tuning} and \emph{transformation}. The transformation module is a generic infrastructure that achieves low-cost layout and loop manipulation by easy-to-use primitive functions. Based on it, the auto-tuning module performs joint data layout and loop optimization by searching in the parameter spaces of the primitive functions. The workflow of \sys is as follows.


%
First, the user provides the computational graph of a deep model, which a domain-specific language (\textit{e.g.}, a subset of Python) can express. It can also be constructed from a model file generated by other frameworks (\textit{e.g.}, TensorFlow \cite{abadi2016tensorflow}). 

Second, the auto-tuning module builds search space for tensors and operators and explores the space jointly. To reduce the tuning time, it uses a cost model to minimize time-consuming on-device measurements.
When the exploration completes, it decodes the best performant point found in the space into a sequence of layout and loop primitives. Then, it delivers these primitives to the transformation module.

Third, the layout propagation submodule propagates layout primitives. Then, the transformation module applies all primitives to perform layout and loop transformation to generate an optimized tensor program. Finally, we deploy the program on different hardware for inference.

\presec
\section{Transformation}

We first introduce the transformation module of \sys, which is a generic infrastructure for manipulating data layouts and loops. The transformation module consists of three submodules: layout transformation, layout propagation, and loop transformation. 

\subsection{Layout Transformation}

\begin{table*}
	\centering
	\caption{Basic layout primitives.}
	\label{tab:basic_prims}
	\begin{tabular}{llll}
		\toprule
		\textbf{Primitive} &\textbf{Parameter} &\textbf{Transformed Shape} & \textbf{Transformed Accessing Expressions}\\
		\midrule
		split & $k$, $F_1, ..., F_m$ &  $...N_{k-1}F_1...F_m N_{k+1}...$& $..., i_{k-1}, \frac{i_k}{F_{2\rightarrow m}}, ..., \frac{i_k}{F_m}\,\, \text{mod}\,\, F_{m-1}, i_k\,\, \text{mod}\,\, F_m, i_{k+1}, ...$\\
		\midrule
		reorder & permutation vector $p$ & 
		$N_{p(1)}N_{p(2)}...N_{p(k)}$ & $i_{p(1)}, i_{p(2)}, ..., i_{p(k)}$\\
		\midrule
		fuse & $k, k+1,..., k+m$ &  $...N_{k-1}(N_{k\rightarrow k+m})N_{k+m+1}...$ & $..., i_{k-1}, (i_{k}N_{2\rightarrow m} + i_{k+1}N_{3\rightarrow m} + ... + i_{k+m}), i_{k+m+1}, ...$\\
		\bottomrule
	\end{tabular}
\end{table*}

To achieve low-cost layout manipulation and easy layout tuning,
we devise various primitive functions to transform data layouts: \emph{split}, \emph{reorder}, \emph{fuse}, \emph{unfold}, \emph{pad}, and \emph{store\_at}. Among them, \textit{split}, \textit{reorder}, and \textit{fuse} are basic primitives and the others are advanced primitives. These primitives lift the data layout transformation from the black-box compiler level to the source level to facilitate leaner control with domain-specific knowledge. We will temporarily cache the operation each time a primitive is applied on a tensor. During program generation, as a compilation pass, we will actually transform the data shapes and alter the corresponding accessing statements in the program. Thus, no human interference is required for re-implementing the operators.

\subsubsection{Basic Layout Primitives}
\label{subsec:basic_prims}
The basic primitives perform one-to-one transformations. 
Given an $n$ dimensional tensor $T$ with original data layout of $N_1N_2...N_n$ and accessing expressions of $i_1, i_2, ..., i_n$, we summarize basic primitives in \cref{tab:basic_prims}, where $1 \leq k \leq n$ is an index to dimensions, $F_k$ is an integer denoting the splitting factor, $p$ is a permutation vector with $p(k)$ as its $k$-th element, and $F_{2\rightarrow m}$ is an abbreviation for $\prod_{i=2}^{m}{F_i}$ ($N_{k \rightarrow k+m}$ is similar). 

For instance, to get the $N\frac{O}{o_t}HWo_t$ layout from $NOHW$, we can apply the following primitive sequence: 
\begin{minted}{python}
split(T, dim=2, factors=[O // o_t, o_t])
reorder(T, perm=[1, 2, 4, 5, 3])
\end{minted}
Alternatively, to pack the layout into spatial blocks, we can transform $NHWO$ through another primitive sequence: 
\begin{minted}{python}
fuse(T, dims=[2, 3, 4])
split(T, dim=2, factors=[O // 4, 4, H * W])
reorder(T, perm=[1, 2, 4, 3])
\end{minted}
During program generation, the first fuse primitive produces shape $N(HWO)$, the second gives $N(\frac{O}{4})4(HW)$, and the final reorder generates $N(\frac{O}{4})(HW)4$, based on \cref{tab:basic_prims}. Assuming the original accessing statement $T[n][h][w][o]$ in the code, it will be transformed as follows:
\begin{enumerate}
	\item T[n][$h(WO) + wO + o$], and let $e = h(WO) + wO + o$
	\item T[n][$\frac{e}{HW4}$][$\frac{e}{(HW)}\,\,\text{mod}\,\,4$][$e\,\,\text{mod}\,\,(HW)$]
	\item T[n][$\frac{e}{HW4}$][$e\,\,\text{mod}\,\,(HW)$][$\frac{e}{HW}\,\,\text{mod}\,\,4$]\, .
\end{enumerate}

\subsubsection{Advanced Layout Primitives}

\label{subsec:adv_prims}
The above examples show the versatility of basic primitives. 
%
%
However, there are cases that cannot be covered, such as the overlapped tiling in \cref{fig:motiv_layout}.
%
%
%
To achieve such special transformations, we abstract advanced layout primitives: unfold, pad, and store\_at.

\textbf{unfold:} This primitive performs overlapped tiling. It accepts a tile\_size parameter, and a stride parameter which is the interval between two tiles: 
\mint{python}|unfold(tensor, dimension, tile_size, stride)|

We denote $tile\_size$ as $B$ and $stride$ as $S$. 
If the original size for a dimension is $D$, this primitive will generate two new dimensions with sizes of $\left(\ceil{\frac{D-B}{S}} + 1\right) $ and $B$. 
For instance, an array $\{1, 2, 3, 4, 5\}$ can be unfolded to a 2-D array $\{\{1, 2, 3\}, \{3, 4, 5\}\}$ by setting $B=3$ and $S=2$. For the input tensor layout in \cref{fig:motiv_layout}, we can set $B = \frac{H}{2} + (KH - 1)$, $S = \frac{H}{2}$ for the height dimension, and the width is similar.
%


The unfold primitive is useful for sliding-window computational patterns, \textit{e.g.}, convolutional layers. 
They have the memory access pattern of $Vi + r$,
where $V$ is the constant convolutional stride, $i$ is the window index, and $r$ is a reduction iterator for the offset inside a window. In the following, we use $M$ to denote the window size (\textit{e.g.}, $M$ will be equal to $KH$ and $KW$ for the two patterns $ih + rh$ and $iw + rw$ in \cref{fig:motiv_program}, respectively). 
%
Then, the original accessing statement $T[Vi + r]$ will be transformed to
 \begin{align}
 T\!\left[\left\lfloor{\!\frac{i}{\floor{\frac{B-M}{V}} \!+\! 1}}\right\rfloor\!\right]\!\left[Vi \!+ \!r \!-\! S\!\left\lfloor{\!\frac{i}{\floor{\frac{B-M}{V}}\!+ \!1}}\right\rfloor\!\right] \,.
 \label{eq:unfold_index_transform}
 \end{align}
Besides unfold, we also propose \textbf{pad} and \textbf{store\_at} primitives.
The pad primitive is to append zeros for a selected dimension, which is useful to align data in memory and alleviate bank conflicts on the shared memory of the NVIDIA GPU. 
The store\_at primitive allows fusing two tensors together by attaching one to another to improve inter-tensor data locality. 
For example, in a fully connected layer, it can attach each element of the bias vector to each column of the weight matrix. 
Subsequently, the inner product and the bias addition in \emph{general matrix multiplication} (GMM) may be computed together by accessing the weight column and the bias element in the same cache line. Additionally, all three primitives have their inverse counterparts, namely {fold}, {unpad}, and {decouple\_at}, to transform layouts back and forth.
\subsection{Layout Propagation}
\label{subsec:layout_prop}


\begin{figure}
	\centering
	\subfloat[Layout conversion operator.]{\label{subfig:layout_conversion_op}{\includegraphics[width=0.36\textwidth]{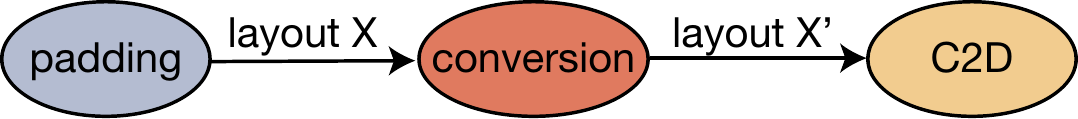}}}
    \\
    \centering
    \subfloat[Layout propagation.]{\label{subfig:layout_prop}{\includegraphics[width=0.23\textwidth]{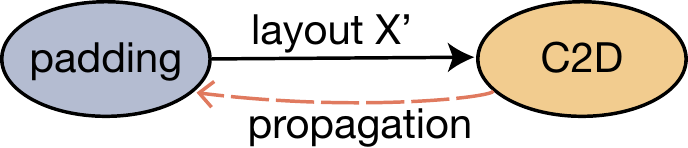}}}
	\label{fig:layout_conversion}
	\vspace{-0.1in}
	\caption{Ways to achieve runtime layout conversion.}
	\vspace{-0.1in}
\end{figure}


The layout primitives working at the local tensor level could incur overhead when performing joint or end-to-end optimization on a computational graph. 
Specifically, we discover two types of such overhead: layout-conversion overhead and fusion-conflict overhead. In this subsection, we will analyze the overhead and propose the layout propagation mechanism to address this issue.


Given a C2D, if it requires a different layout for the weight tensor, we can transform it offline without any runtime overhead because the weight tensor is a constant. Unfortunately, if the C2D requests a different input layout $\mathcal{X^\prime}$, it can only be achieved either by (1) inserting an operator performing runtime layout conversion (\cref{subfig:layout_conversion_op}) or (2) letting the producer operator yield each element based on the new layout directly (\cref{subfig:layout_prop}). Inserting layout conversion operators will incur extra overhead due to runtime data movements. 
So, we prefer the second way, which is called {layout propagation}. After propagation, the padding operator actually performs two tasks at runtime: padding zeros and converting the layout.
Similarly, for the output tensor of C2D, we can let its consumer operator access the new layout directly, rather than inserting another conversion operator next to C2D. 


\begin{figure}
     \centering
     \begin{minted}[fontsize=\small]{python}
for n in range(N):
  for ht in range(H // 4):
    for w, o in range(W, O):
      for hi in range(4):
        Conv[n][ht][w][o][hi] = 0.0
        for ri, rh, rw in range(I, KH, KW):
          Conv[n][ht][w][o][hi] += Inp[...]*Ker[...]
for n, o, h, w in range(N, O, H, W):
  ReLU[n][o][h][w] = max(Conv[n][h//4][w][o][h%4],0)
\end{minted}
\vspace{-0.08in}
     \caption{Loop nests without propagation and fusion.}
     \vspace{-0.15in}
     \label{fig:loop_wo_prop}
\end{figure}

Besides the layout-conversion overhead, another delicate issue emerges when incorporating operator fusion.
Operator fusion is a loop-tuning technique to promote inter-operator data locality by letting the downstream operator consume the intermediate data immediately before spilling out of the cache. 
Consider two operators: C2D and ReLU, and
%
%
the original output layouts of them are both $NOHW$. Suppose we transform the output layout of the C2D to $N\frac{H}{4}WO4$ through split and reorder primitives. 
Then, the generated program is shown in \cref{fig:loop_wo_prop}. The loop nest of the C2D is reconstructed accordingly due to the output layout transformation.
Different from the original case, we cannot perform loop tiling on the two loop nests with the same tile sizes and then fuse the two nests. Since fusion is an effective technique, reducing the chance of fusion due to the reconstructed loop nest will result in performance loss.
%


\begin{figure}
     \centering
     \begin{minted}[fontsize=\small]{python}
for n, ht, w, o, hi in range(N, H // 4, W, O, 4):
  Conv[n][ht][w][o][hi] = 0.0
  for ri, rh, rw in range(I, KH, KW): 
    Conv[n][ht][w][o][hi] += Inp[...] * Ker[...]
  ReLU[n][ht][w][o][hi] = max(Conv[...], 0)
\end{minted}
     \vspace{-0.06in}
     \caption{Loop nests with propagation and fusion.}
     \vspace{-0.15in}
     \label{fig:loop_w_prop}
\end{figure}

To eliminate such fusion-conflict overhead induced by layout transformation, we extend the layout propagation mechanism such that the same layout can be shared among multiple tensors. Layout propagation can be implemented easily by duplicating the primitive sequence of the source tensor for the target tensor. %
For instance, we replicate the primitives from tensor $Conv$ in \cref{fig:loop_wo_prop}, \textit{i.e.}, split and reorder primitives, for tensor $ReLU$. Then ReLU will trigger the same loop nest reconstruction, hence aligned perfectly with that of C2D. Consequently, the fusion-after-tiling in loop tuning will be the same as the normal case, as illustrated in \cref{fig:loop_w_prop}. 

Although layout propagation helps to eliminate the overhead incurred by layout transformation, it has three constraints. First, we only propagate primitives along a path with only element-wise operators and among tensors with the same shape.
Given an operator $Y[i] = F(X[i])$, there exists an element-wise data mapping between the output tensor $Y$ and the input tensor $X$. 
We can propagate the layout of $Y$ to $X$, or vice versa.
This constraint is introduced because the parameters of primitives are shape-dependent. Second, we \emph{will not} propagate a primitive sequence if it contains non-trivial advanced primitives. This is because advanced primitives will induce data expansion. Instead, 
%
we will insert conversion operators when they arise, as in \cref{subfig:layout_conversion_op}. 
Third, the layout tuning for each complex operator will be performed independently. This constraint is to eschew the overhead of layout propagation itself, because the optimal layout of a complex operator may lead to inferior performance for another. For example, given two consecutive C2Ds, we will insert a conversion operator between them if needed rather than letting the output tensor of the former C2D and the input tensor of the latter C2D share the same layout. Notably, no conversion operator is necessary when other simple operators exist between the two C2Ds (\textit{e.g.}, we can propagate a layout onto the padding operator like in \cref{subfig:layout_prop} and let it perform the actual layout conversion).
\subsection{Loop Transformation}
\label{subsec:loop_trans}


We perform loop transformation via reusing the loop primitives of TVM \cite{chen2018tvm}: split, reorder (same names as layout ones, but distinct functions), vectorize, unroll, cache\_read/write, parallel, inline, and compute\_at. 
%
Most loop-tuning techniques, including loop tiling, vectorization, and operator fusion, can be realized by combining these primitives.

\presec
\section{Auto-tuning}
\label{sec:automation}

Even with the transformation module, 
optimization is still painful because it requires numerous manual trials. The combination of layout and loop tuning further exacerbates the problem.
Thus, in the auto-tuning module, we devise a unified framework to jointly optimize layouts and loops to generate high-performance programs automatically. 

Our joint tuning comprehends three steps: 1) we build the layout tuning space for tensors and loop tuning space for operators, each point in the space can be decoded as a primitive sequence; 2) we explore the tuning space to find the best performant point; 3) we decode this point as instantiated primitives and deliver them to the transformation module. 

\subsection{Space Building}
\label{subsec:space_build}
Auto-tuning is to search for the code with the best performance in the tuning space. With our transformation module, we only need to find the best parameters to apply primitives. Thus, the tuning space is equivalent to the parameter space for primitives. For now, we only consider layout split, reorder, and unfold primitives in the layout space. Also, we will omit details on the loop space, which is similar to \cite{zheng2020flextensor, zheng2020ansor}, \textit{e.g.}, space of loop split factors for each operator.

The layout space to be built should be pruned, otherwise, it will be infinitely large because the number of primitives that can be applied is infinite. As in \cref{sec:intro}, we only perform layout tuning for complex operators and propagate their results to reduce the number of tuning tasks. Further, we craft a layout tuning template for each tensor that is accessed by complex operators. Each template only exposes a subset of parameters of primitives as tunable options. The templates are crafted based on the following observations on how data layouts influence performance considering intra-operator data dependency and hardware characteristics.

First, data layout influences the data reuse strategy, \cite{kandemir1998enhancing, clauss2000automatic, lu2009data, miucin2018data}. For most architectures, data reuse is vital to reducing the number of memory accesses and improving software pipeline. Consider the C2D as an example, each output element requires $(KH) \cdot (KW) \cdot I$ input elements for reduction. Without data reuse, we need totally $N\cdot H \cdot W \cdot O \cdot (KH) \cdot (KW) \cdot I$ load instructions for the input tensor. Fortunately, an input element is required by at most $(KH) \cdot (KW) \cdot O$ output elements. Thus, we can reuse an input element to accumulate on $KH\times KW$ spatial positions or $O$ channels before spilling it out of the cache. Besides, sequential data accesses can be bundled by SIMD instructions. With these two aspects, we can also explain why $NHWO$ layout often performs better than $NOHW$ layout \cite{zheng2020ansor}: 1) an input element can be reused to accumulate on many (at most $O$) output channels and $O$ is typically large, hence a high reuse rate; 2) output channels can be loaded with SIMD instructions easily since $O$ is the last dimension. 

\begin{table}
	\centering
	\small
	\caption{Profiled L1 data cache misses.}
	\begin{tabular}{lcc}
		\toprule
		\textbf{Tile Size} &\textbf{\#L1-mis / Pred. (1st F.)} &\textbf{\#L1-mis (2nd F.)}\\
		\midrule
		$512\times 4$ & 32 / 32 & 208 \\
		\midrule
		$512 \times 16$ & 96 / 128 & 262 \\
		\midrule
 		$512 \times 64$ & 501 / 512 & 785 \\
 		\midrule
 		$512 \times 256$ & 2037 / 2048 & 2952 \\
		\bottomrule
	\end{tabular}
	\label{tab:cache_prefetch_prof}
\end{table}

Second, data layout influences cache utilization. Both layout and loop tiling can be exploited to let a data block fit in cache \cite{shirako2020affine}. Besides, we also observe that layout tiling can further prevent cache misses by facilitating hardware prefetching \cite{chilimbi1999cache, monil2020understanding, cronin2021exploration}. To verify this, we conduct an experiment on a Cortex-A76 CPU, which is a big core on Kirin 990 SoC, the L1 data cache line size of which is float32x16 (\textit{i.e.}, 64 bytes). We profile two functions and both of which only load a 2-D data block once from memory with NEON instructions. The data elements for the first function are stored contiguously in memory, \textit{i.e.}, layout tiling case. By contrast, the elements for the second function are stored row by row, 
\textit{i.e.}, loop tiling case without changing data placements. The profiled L1 cache misses are reported in \cref{tab:cache_prefetch_prof}, where we also present our predictions based on hardware prefetching in the second column. We observe that the CPU is very likely to fetch four contiguous cache lines when a miss event is triggered. For example, the prediction for tile size $512\times4$ is calculated as $\frac{512 \times 4}{16 \times 4} = 32$. From \cref{tab:cache_prefetch_prof}, layout tiling is preferable to loop tiling to improve cache utilization via hardware prefetching. Most importantly, the cache performance after layout tiling is always better than in other cases.


The second observation indicates that layout tiling improves cache utilization even though loop tiling has been exploited. Thus, our layout tuning template is a tiling template, with tiling sizes as basic tunable options. For most dimensions, the tiling can be achieved with split primitives. For height and width dimensions of convolutions, it can be achieved with the unfold primitives to enable the overlapped tiling. After splits and unfolds, based on the first observation, we let the tiled channel dimension be the last dimension to promote data reuse and SIMD. Consequently, our data layout tuning template for C2D has the following form:

\begin{itemize}[leftmargin=*]
    \item output tensor $Conv$: $N\frac{H}{h_t}\frac{W}{w_t}\frac{O}{o_t}h_t w_t o_t$, where $h_t$, $w_t$, and $o_t$ are \emph{three tunable} split parameters for tiling $H$, $W$, and $O$.
    \item input tensor $Inp$: $N\frac{H}{h_t}\frac{W}{w_t} \frac{I}{i_t} \left(h_t + KH-1\right) \left(w_t + KW - 1\right) i_t$, where $\left(h_t + KH - 1\right)$ and $\left(w_t + KW - 1\right)$ are the unfolded dimensions, and $i_t$ is the \emph{only tunable} split parameter for tiling $I$. 
    \item weight tensor $Ker$: $\frac{O}{o^\prime_t}\frac{I}{i^\prime_t}(KH)(KW)i^\prime_t o^\prime_t$, where $i^\prime_t$ and $o^\prime_t$ are \emph{two tunable} split parameters for tiling $I$ and $O$.
\end{itemize}

In the above templates, uppercase letters represent the original dimensions, while lowercase letters with a subscript $t$ denote the tiled parameters correspondingly. We do not need to tune the unfolded dimensions for the input tensor, because they are directly related to the tiling of the output tensor. Suppose the tuner splits the $H$ dimension of the output tensor as $\frac{H}{h_t} \times h_t$. It then applies the following unfold primitive on the input tensor directly:
\begin{minted}{python}
unfold(Inp, Inp height, h_t + (KH - 1), h_t)
\end{minted}
This is the same as the case in \cref{fig:motiv_layout} where $h_t = \frac{H}{2}$. 

In summary, the pruned layout space for C2D consists of six tunable parameters (\textit{i.e.}, at a scale of $O(10^6)$): $h_t, w_t, o_t$ for tiling $H, W, O$ of the output tensor, $i_t$ for tiling $I$ of the input tensor, $i_t^\prime, o_t^\prime$ for tiling $I, O$ of the weight tensor. For other convolutions (\textit{e.g.}, 3-D case), the template is similar. 

For a GMM $C = A \bigodot B$, where $MN, MK, KN$ are the original layouts of the three matrices, the search space is much smaller due to fewer dimensions. Thus our template consists of split parameters for all dimensions. Then, based on the first observation, the reorder after splits is determined without tuning: $\frac{M}{m_t}\frac{N}{n_t}m_t n_t$ for $C$, $\frac{M}{m_t}\frac{K}{k_t}m_t k_t$ for $A$, and $\frac{K}{k_t}\frac{N}{n_t}k_t n_t$ for $B$. Finally, there are three tunable parameters (\textit{i.e.}, up to $O(10^3)$ points): $m_t, k_t, n_t$, in the layout space for GMM.

The above templates only perform one-level multi-dimen-sional layout tiling. We can expand them to multi-level cases easily, which can be configured in \sys for scalability. For example, we can use two-level layout tiling templates for \sys, where the template for the output tensor of C2D can be defined as $N\frac{H}{h_{t}^\prime h_t}\frac{W}{w_{t}^\prime w_t}\frac{O}{o_{t}^\prime o_t}{h_{t}^\prime w_{t}^\prime o_{t}^\prime }h_t w_t o_t$. 

Without our template-based pruning, the search space, especially the parameter space for the reorder primitive, will be too large to explore. The only concern after pruning is whether the subspace contains good points. We verify the effectiveness of pruning through experiments. 

\subsection{Exploration \& Cost Model}
\vspace{0.02in}
\label{subsec:exp_eval}

\begin{figure}[t]
	\centering
	\includegraphics[width=0.4\textwidth]{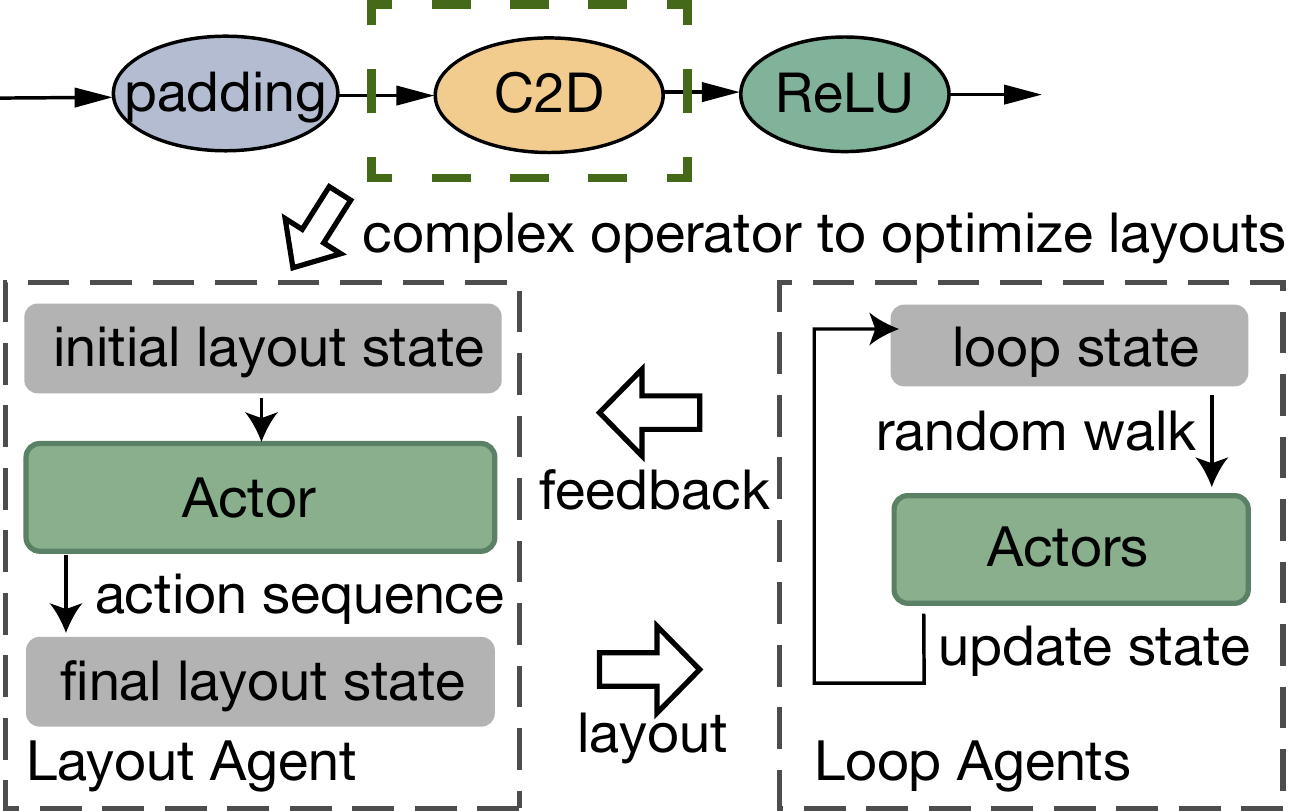}
	\vspace{-0.05in}
	\caption{Cross exploration architecture.}
	\vspace{-0.15in}
	\label{fig:exp_arc}
\end{figure}

To explore the search space, we need to: (1) visit points efficiently; (2) evaluate visited points rapidly. We resort to the PPO algorithm \cite{schulman2017proximal} from reinforcement learning (RL) to explore the space. Compared with heuristic algorithms (\textit{e.g.}, genetic algorithm) and other RL algorithms, PPO is learning-based and more stable \cite{hernandez2019survey}, which is introduced in \cite{ahn2020chameleon} to speed up the tuning space exploration. To speed up the evaluation, we develop a cost model to predict the performance to reduce the number of time-consuming on-device measurements. 

In RL, an \emph{agent} will respond (referred to as \emph{action}) to environments based on its \emph{observation}, which is composed of the \emph{state} of the current environment and feedback given by the environment called \emph{reward}. 
PPO employs two neural networks: \emph{actor} and \emph{critic}. The actor gives actions while the critic judges each action, \textit{i.e.}, fitting the real rewards.

Even with PPO, exploring layout and loop spaces simultaneously is challenging. Consider the C2D as an example, we need to rebuild its loop space every time given a new layout, because the loop nest relies on the output layout, like $n, o, h, w$ in \cref{fig:loop_wo_prop}. The reconstructed loop space further leads to that the points searched previously will be invalid in the new space, hence inefficient exploration. 

As in \cref{sec:intro}, our solution to this issue involves two aspects. We first divide the performance tuning into two stages: the joint stage and the loop-only stage. We then propose a cross-exploration architecture, as shown in \cref{fig:exp_arc}, for the joint stage. The cross-exploration repeats the following process: determining a layout through the layout PPO actor, performing multiple rounds of loop tuning via loop PPO actors, and feeding the best performance back as the reward for the current layout. Consequently, we achieve a bidirectional and unified optimization flow in the joint stage to find better layouts. We also prevent inefficient loop tuning, since the loop reconstruction will not occur in the loop-only stage.

In the following, we will only elaborate on the design of RL action, state, and reward for the joint stage based on the cross-exploration architecture. The loop-only stage can be achieved by removing layout-related searches.

\subsubsection{Layout Space Exploration} 

Since the pruned layout space only involves tunable split parameters, we here develop a generic actor to explore the parameter space of the layout split primitive. Then, the final layout will be determined by a sequence of actions. Take the C2D in \cref{fig:loop_wo_prop} as an example, the action sequence for resolving the output layout of $Conv$ consists of: split $H$, split $W$, split $O$, and reorder them to $N\frac{H}{h_t} \frac{W}{w_t} \frac{O}{o_t} h_t w_t o_t$. The split actor only provides the factors to split $H, W, O$, while the reorder is determined in the template in \cref{subsec:space_build}. Similarly, replacing the first two splits with unfolds forms the action sequence for the input layout.

Consider the dimension with a size of $D$ in a tensor.
%
To obtain a generic split actor, we map its output action $a_s$ to a contiguous interval $(0, 1)$. Then, the splitting factor $F$ is calculated as follows:
\begin{equation}
    F = R(D \cdot a_s) \, .
\end{equation} Assume the tensor $Conv$ in \cref{fig:loop_wo_prop} has $O = 32$. The actor gives one action $a_s = 0.5$. Then we derive two split dimensions : $o_t = R(32 * 0.5)=16, \frac{O}{o_t} = R(32 / 16)=2$. 

The state for the actor is given by the \emph{concatenation} of the current states of all primitives for all tensors of the complex operator (\textit{e.g.}, $Inp$, $Ker$, $Conv$ in a C2D). For instance, when unfolding the height of $Inp$ in \cref{fig:motiv_layout} into two parts, the current state of the unfold primitive is changed to $[2, \frac{H}{2} + (KH - 1)]$, while the initial state was $[1, H + KH - 1]$. Similarly, the current state for the split primitive is composed of factors, \textit{e.g.}, $[2, 16]$ for $o=32$ (initial state was $[1, 32]$). Then the final state is the concatenation of all such sub-states.

\subsubsection{Loop Space Exploration} 
The exploration for loop space follows a similar random-walk design as \cite{zheng2020flextensor}. We first sample a \emph{batch} of points in the loop space and choose the best one as the starting point, then each actor gives a direction for some parameter space. After that, we arrive at the next point by walking along that direction, as shown in \cref{fig:exp_arc}. 

Including the layout split actor, we have a lot of actors now. 
To model the interference among subspaces/primitives, we deploy a \emph{global shared critic network} for all actors (not shown in \cref{fig:exp_arc} for simplicity). 

The reward $r$ for all RL agents is the same:
\begin{align}
	r = U - l \, ,
	\label{eq:reward_func}
\end{align} where $U$ is a constant and $l$ is the latency of some point. For layout RL agents, $l$ is chosen as the best latency after several rounds of loop exploration given the current layout.

\subsubsection{Cost Model} 

To evaluate points rapidly, we estimate the performance by developing a cost model for each hardware. The cost model is a tree ensemble from XGBoost \cite{chen2016xgboost}, similar to that of Ansor \cite{zheng2020ansor}. 
For some point, we decode it as primitives and apply them to generate the optimized tensor program. Then we feed the features of the program (\textit{e.g.}, loop structures and accessing expressions) to the cost model to estimate the throughput. 
During exploration, we only measure the top-$k$ points of a batch or an episode of RL trajectories, which are predicted by the cost model, on the target hardware. These measurements are also used for training the cost model online.

\begin{figure*}
	\centering
	\subfloat[Single Operator on Intel CPU.]{\label{subfig:exprim_op_llvm}{\includegraphics[width=0.33\textwidth]{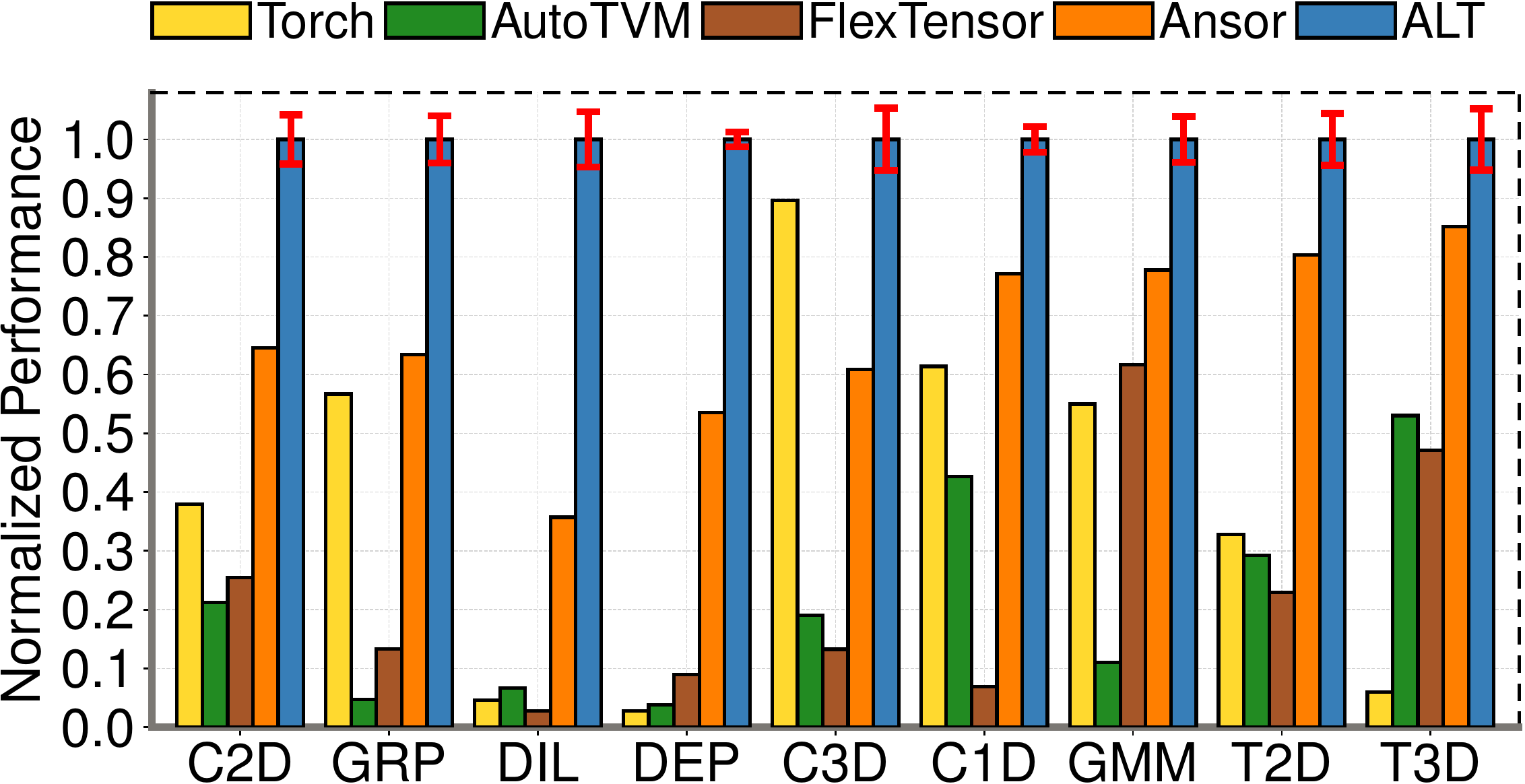}}}\hfill
	\subfloat[Single Operator on NVIDIA GPU.]{\label{subfig:exprim_op_cuda}{\includegraphics[width=0.33\textwidth]{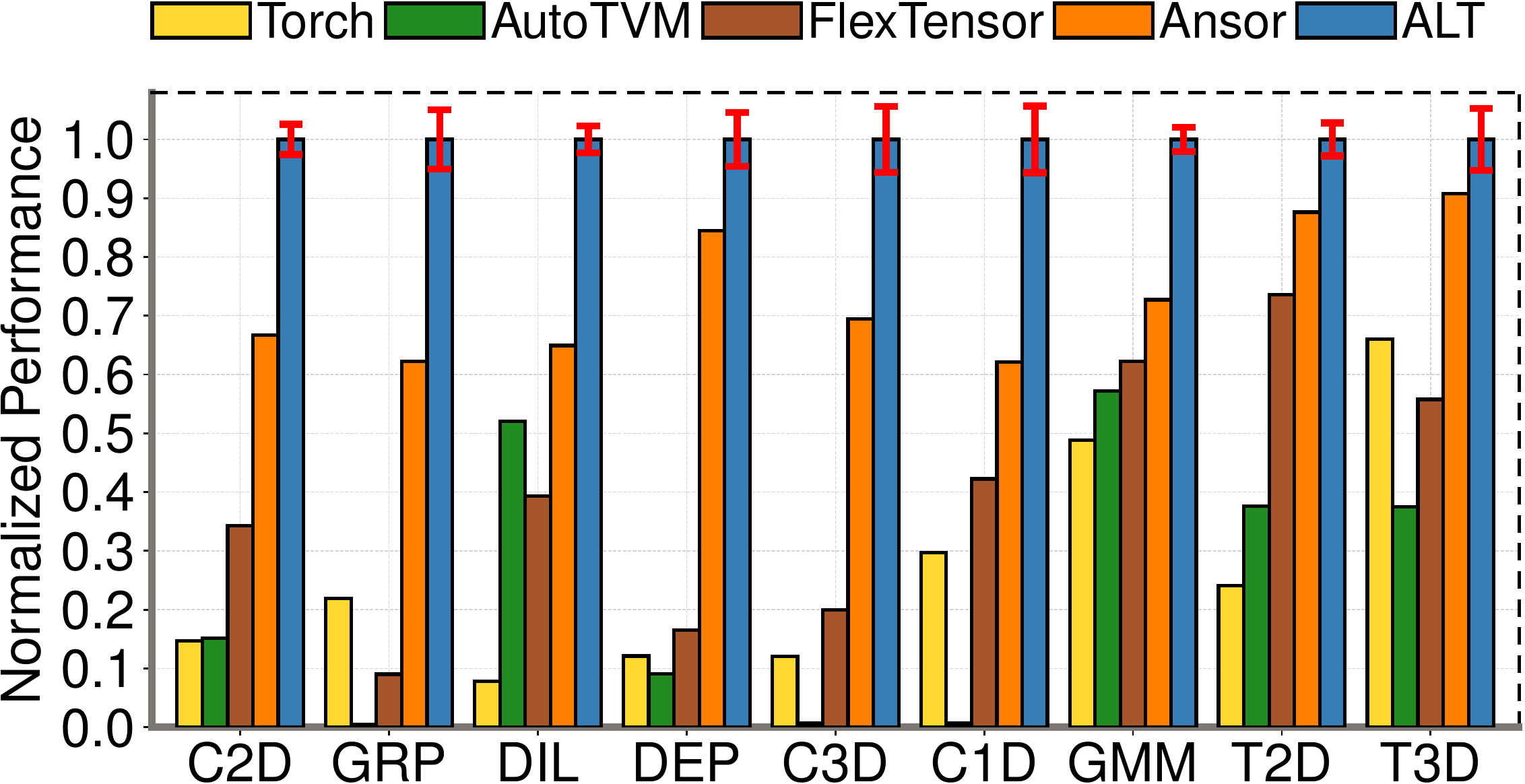}}}\hfill
	\subfloat[Single Operator on ARM CPU.]{\label{subfig:exprim_op_android}{\includegraphics[width=0.33\textwidth]{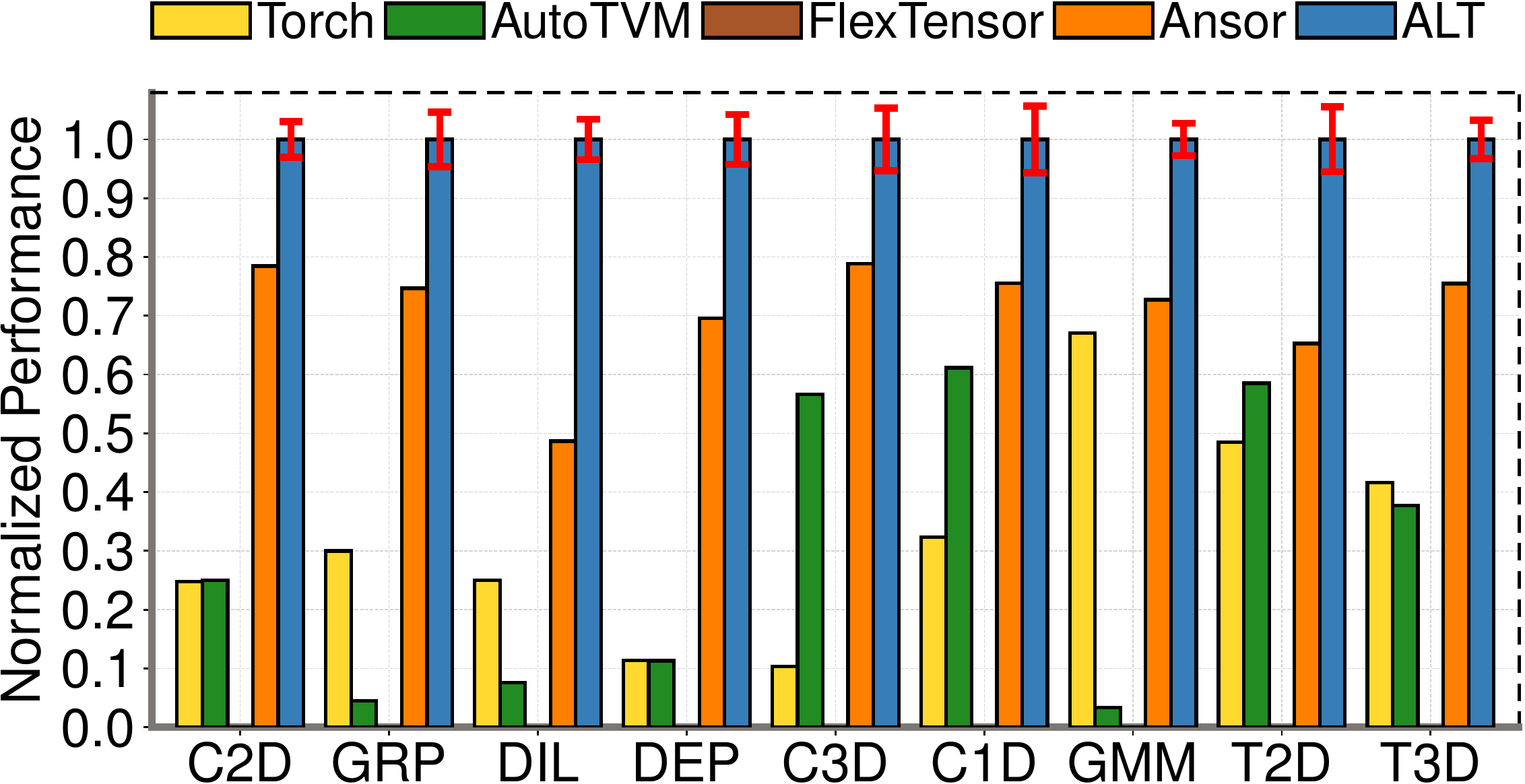}}}\hfill
	\vspace{-0.05in}
	\caption{Single operator performance.}
	\label{fig:exprim_op}
	\vspace{-0.13in}
\end{figure*}

\vspace{-0.05in}\section{Implementation}
\label{sec:impl}
We implemented \sys based on TVM (v0.8dev1) \cite{chen2018tvm} with 19K LoC of Python and 2K LoC of C++. 

To implement the layout transformation, {we insert a pass before lowering the tensor expression (TE) of TVM to TVMIR. This pass} will rewrite the indices of all tensor accesses in TE when layouts change. With regard to an operator $Y = F(X)$ where the output tensor $Y$ is of shape $N_1N_2..N_m$, in TE this operator has $m$ nested
spatial loops, each corresponding to a dimension of $Y$ (one-to-one mapping). We denote the loop variables as $L=\{l_1, l_2, ..., l_m\}$. Assume \sys caches a set of primitive sequences $\mathcal{S}$ either provided manually or by the auto-tuning module automatically. Our pass will first transform accesses for the output tensor $Y$, and then other tensors. We denote the primitive sequence for $Y$ as $\mathcal{S}(Y)$ (abbreviated as $S_Y$). Our pass first deducts the final layout of $Y$ by applying each primitive function in $S_Y$. Assuming the new layout has $n$ dimensions with shape $N_1^\prime N_2^\prime..N_n^\prime$, the loop structure will then be reconstructed by TE as $L^\prime = \{l_1^\prime, l_2^\prime, ..., l_n^\prime\}$.
{Given the one-to-one mapping between a dimension of the output tensor and a loop variable, we will also have $L^\prime = S_Y(L)$.} With this, we can transform accesses for tensor $X$ while ensuring validity. Specifically, the accesses of $X$ must first be remapped with the newly reconstructed loop variables. The remapping is done in two steps: 1) calculating the inverse primitive sequence of $S_Y$, denoted as $S_Y^{-1}$; 2) replacing all old loop variables $L$ by $S_Y^{-1}(L^\prime)$ in all access indices of $X$. After this remapping, the tensor accesses of $X$ can be safely transformed to $S_X(S_Y^{-1}(L^\prime))$ by applying $\mathcal{S}(X)$.

To implement the layout propagation, we copy the primitive sequence of the source tensor for the destination tensor.
The joint stage of \sys sequentially tunes each complex operator following the topological order and propagates the resulting layouts.
A special case is that an operator can have multiple consumers or producers.
In the case of multiple consumers, \sys will propagate the layout of the source tensor to all consumers.
For the case of multiple producers, consider $Y[i] = F(X_0[i], X_1[i], X_2[j])$, where there are element-wise mappings between $X_0$ and $Y$, and between $X_1$ and $Y$.
When the layouts of $X_0$ and $X_1$ are both tuned, \sys will heuristically choose $X_0$ for propagation onto $Y$.
Conversely, if the layout of $Y$ is tuned first (\textit{i.e.}, there is no complex operator prior $X_0$ or $X_1$ that can propagate layouts to them),
\sys will propagate the layout of $Y$ to both $X_0$ and $X_1$.
\presec\section{Evaluation}

%
In this section, we evaluate \sys on various hardware platforms, including 40-core Intel Xeon Gold 6248 CPU@2.5GHz (443GB memory), NVIDIA Tesla V100 (CUDA v11.0), and Kirin 990 SoC (Android v10). We compare \sys with state-of-the-art frameworks and compilers: Torch (v1.7) \cite{paszke2019pytorch}, AutoTVM (v0.8dev1) \cite{chen2018learning}, FlexTensor \cite{zheng2020flextensor}, and Ansor \cite{zheng2020ansor}.
Torch is a reference point for vendor libraries, which was evaluated by using MKL-DNN library \cite{mkldnn} for Intel CPU, cuDNN (v8.0.4) library \cite{chetlur2014cudnn} for NVIDIA GPU, and XNNPACK library \cite{xnnpack} for ARM CPU. AutoTVM, FlexTensor, and Ansor are three widely used auto-tuning frameworks.
Besides, Ansor outperforms Tensorflow Lite \cite{abadi2016tensorflow} and other hardware-specific compilers \cite{liu2019optimizing, zheng2020ansor} such as OpenVINO \cite{openvino} and TensorRT \cite{tensorrt}. Thus, we do not include them as baselines here.

For \sys, if not specified, we use one-level layout tiling templates for layout space building. For loop space exploration, we set the sampling batch size and the episode length to 128, 
and measure the top-8 points predicted by the cost model on the target hardware. In addition, we take the total number of such on-device measurements as a metric of the search budget for all auto-tuning methods. Thus, a batch or an episode of points in \sys will cost a budget of 8.

\subsection{Single Operator Benchmark}
\label{subsec:single_op_bench}

We first present the results on single operators. 
We consider 9 operators, including C2D, Group-wise C2D (GRP), Depth-wise C2D (DEP), Dilated C2D (DIL), 3-D convolution (C3D), 1-D convolution (C1D), GMM, Transposed C2D (T2D), Transposed C3D (T3D). Each operator is evaluated using 10 random configurations with different batch sizes, kernel sizes, etc. For instance, the value of batch size is selected from $[1, 16]$, and the number of input channels is uniformly sampled from $[3, 16, 32, 64, 512, 960, 1280]$. We generate 90 test cases for each device. The result is normalized based on the geometric mean of speedups over the worst latency of each test case. For C1D, C2D/T2D, and C3D/T3D and their variants, we test $NOW/NWO$ for C1D, $NOHW/NHWO$ for C2D/T2D, and $NODHW/NDHWO$ ($D$ is the depth dimension) for C3D/T3D and report the best for baselines except Torch (it only supports $NOW, NOHW, NODHW$). We set the search budget to 1000 for all auto-tuning methods, which is suggested by Ansor. For \sys, the budget for the joint stage and the loop-only stage is 300 and 700 respectively.

As shown in \cref{subfig:exprim_op_llvm}, on Intel CPU \sys achieves $9.5\times$, $9.9\times$, $9.8\times$, and $1.6\times$ speedups in comparison with Torch, AutoTVM, FlexTensor, and Ansor respectively. Among all operators, DIL and DEP have lower operational intensity (the ratio of the number of computational instructions to the number of memory access instructions), and thus they are more likely to be memory-bound. For DIL and DEP, \sys outperforms other baselines with a large margin because layout tuning can effectively reduce memory accessing overheads. Even for operators that are typically compute-bound, \textit{e.g.}, C2D and C3D, \sys still achieves notable speedups. This is because the operational intensity depends on tensor shapes. \sys can tailor the tensor layouts toward each specific shape and hardware platform.

We achieve similar results on NVIDIA GPU and ARM CPU. Compared with Ansor, \sys achieves an average of $1.5\times$ speedup on NVIDIA GPU (\cref{subfig:exprim_op_cuda}), and $1.4\times$ speedup on ARM CPU (\cref{subfig:exprim_op_android}). We do not include the results of FlexTensor for ARM CPU since it does not support ARM backends. Generally, auto-tuning methods can outperform Torch because non-typical operator configurations are often less optimized in vendor libraries. Further, AutoTVM suffers from small tuning space and FlexTensor has no cost model, thus both demonstrate inferior performance than Ansor and \sys. Additionally, compared with Ansor, \sys can effectively tune data layouts with feedback from operator-level optimization and hence illustrate significant improvements.


\presub\subsection{End-to-End Benchmark}
\label{subsec:e2e_bench}
\begin{figure}
	\centering
	\centering
	\subfloat[Network on Intel CPU.]{\label{subfig:exprim_network_llvm}{\includegraphics[width=0.455\textwidth]{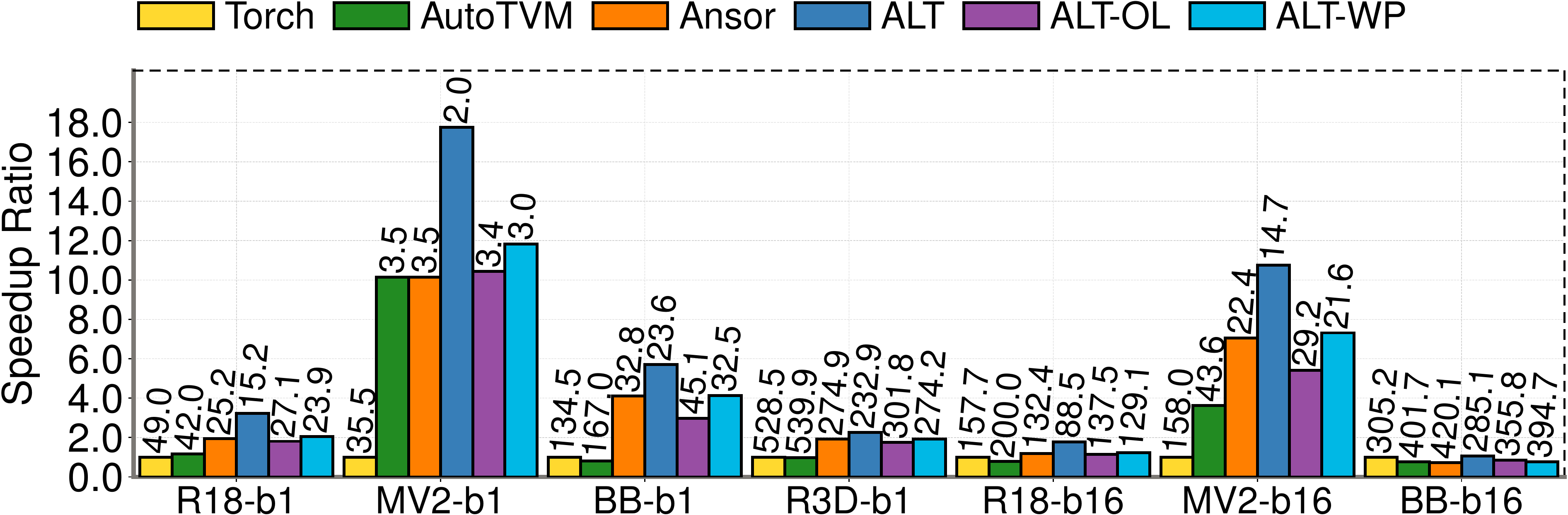}}}\hfill
	\\
	\centering
	\subfloat[Network on NVIDIA GPU.]{\label{subfig:exprim_network_cuda}{\includegraphics[width=0.455\textwidth]{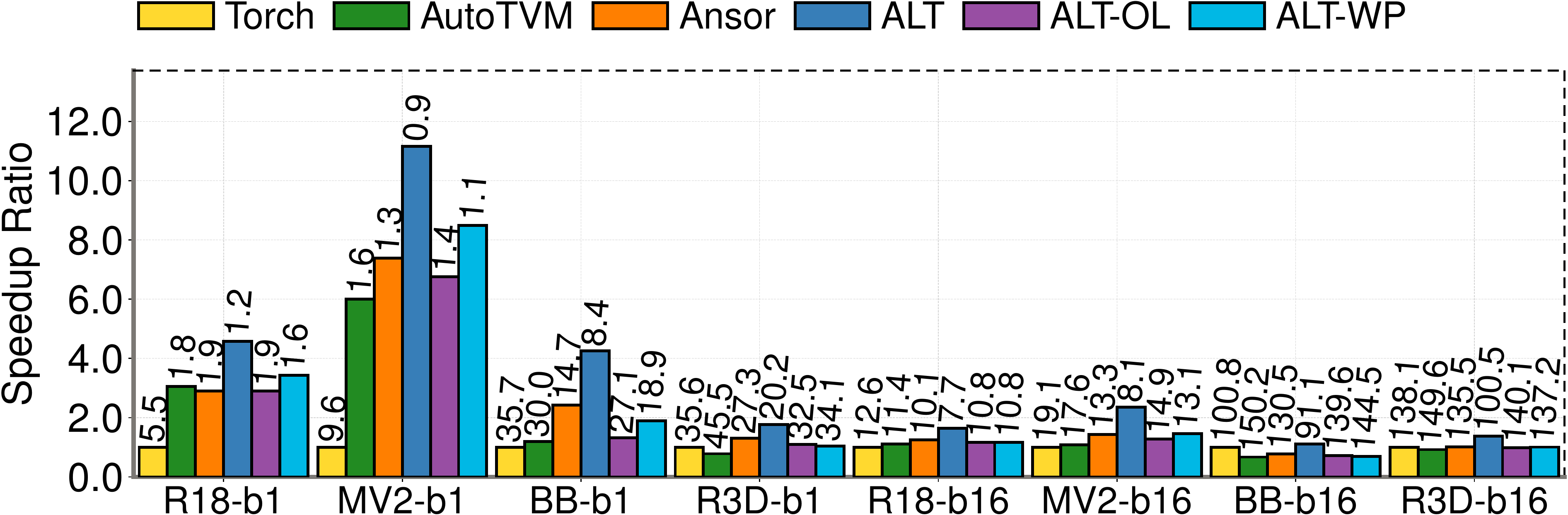}}}\hfill
	\\
	\subfloat[Network on ARM CPU.]{\label{subfig:exprim_network_android}{\includegraphics[width=0.455\textwidth]{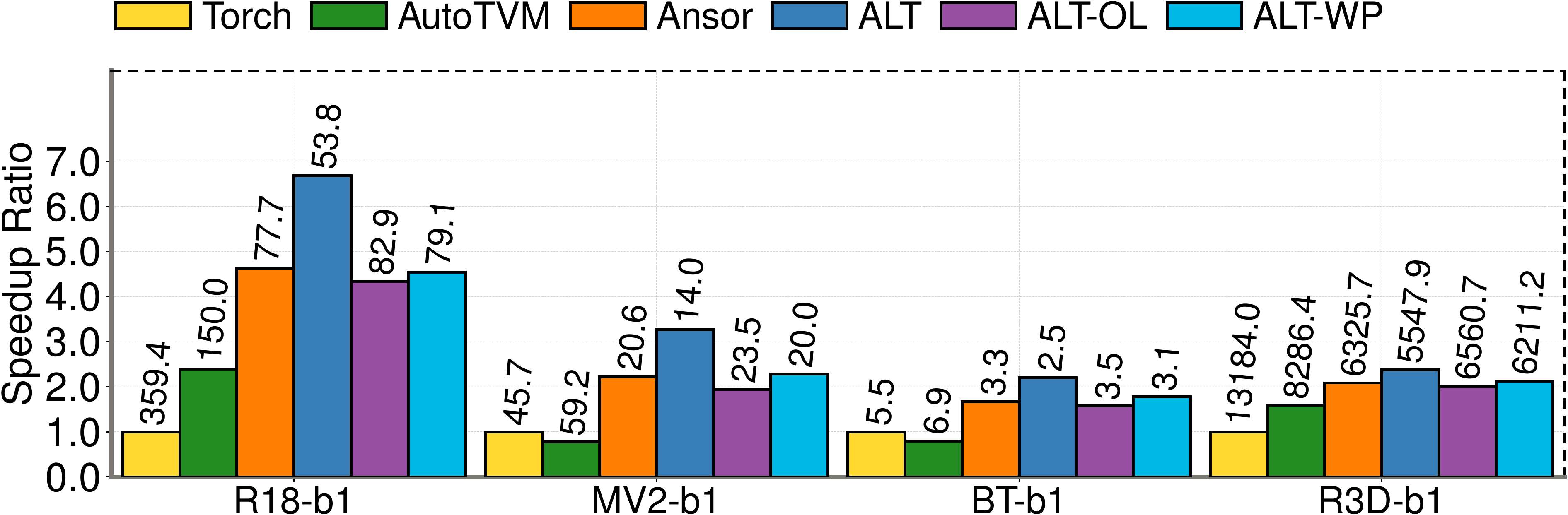}}}\hfill
	\\
	\vspace{-0.05in}
	\caption{End-to-end inference performance.}
	\label{fig:exprim_network}
	\vspace{-0.15in}
\end{figure}
We then evaluate the end-to-end performance of \sys with five neural networks, including applications of 1) image processing: ResNet-18 (R18) \cite{he2016deep}, MobileNet-V2 (MV2) \cite{sandler2018mobilenetv2}, 2) natural language processing: BERT-base (BB) \cite{devlin2018bert}, BERT-tiny (BT) \cite{jiao2019tinybert}, and 3) video processing: ResNet3D-18 (R3D) \cite{hara2017learning}. 
For Intel CPU and NVIDIA GPU, the benchmarks use batch sizes of 1 and 16. For ARM CPU, we set the batch size to 1 due to the limited resource.


For convolutional networks, the input tensor is of shape $N \times 3 \times 224 \times 224$ (image processing) and $N \times 3 \times 16 \times 112 \times 112$ (video processing), respectively. For BERT, the shape of the input tensor is $N \times 128$. For auto-tuning baselines, we set the search budget as 20,000 (which is suggested by Ansor \cite{zheng2020ansor}). We set the budget for the joint stage to 8,000 and the budget for the loop-only stage to 12,000 in \sys. Additionally, Torch uses $NOHW/NODHW$ layouts while AutoTVM and Ansor use $N\frac{O}{o_t}HWo_t/N\frac{O}{o_t}DHWo_t$ after integrating NeoCPU \cite{liu2019optimizing}. 

We illustrate the speedup ratio of all methods over Torch in \cref{fig:exprim_network}, where $b1$ denotes batch size 1 and $b16$ denotes batch size 16. The number on top of each bar demonstrates the latency in milliseconds. To verify the effectiveness of the joint tuning and layout propagation, we define two variants of \sys: (1) \sys-OL, which only involves loop optimization without the joint stage based on $NHWO/NDHWO$ layouts; (2) \sys-WP, which only eliminates conversion operators between adjacent operators, as that shown in \cref{subfig:layout_prop}. Compared with Ansor\footnote{Ansor performs better than Torch \cite{paszke2019pytorch} and AutoTVM \cite{chen2018learning}. We omit the results of Torch and AutoTVM due to the lack of space.}, \sys achieves $1.47\times$, $1.39\times$, and $1.46\times$ speedups on Intel CPU, NVIDIA GPU, and ARM CPU, respectively. For R3D, most of its operators are compute-bound, thus \sys achieves similar results with Ansor. For MV2, which is a lightweight network with lower operational intensity, \sys outperforms the baselines significantly. 


Notice that \sys-OL achieves similar performance as Ansor because both of them mainly involve loop tuning. When incorporating layout tuning and basic layout propagation, \sys-WP shows $1.1\times$ speedup over \sys-OL in general and no improvement in a few cases. \sys achieves $1.3\times$ speedup on average compared with \sys-WP. This is because operator fusion is a critical loop-tuning technique to improve performance, while \sys-WP cannot combine layout tuning and loop tuning effectively. 

\presub\subsection{{Micro Benchmark}}
\label{subsec:micro_bench}

We dive into the details to achieve a better understanding of the system design. First, we present the overhead of layout propagation.
We then study the parameter sensitivity of \sys in the end-to-end optimization.
%
%
We will also conduct a case study to help understand the searched layouts and loops. 
Finally, we present more observations to provide hints for deep compiler optimization. 
Notably, we do not give more experiments on cost model \cite{zheng2020ansor} and the PPO exploration method \cite{ahn2020chameleon} because they are not our major contributions. 

\subsubsection{Layout Propagation Overhead:} 
We here study the overhead of layout propagation to show the necessity of the introduced constraints in \cref{subsec:layout_prop}. We evaluate two subgraphs on 48-core Intel(R) Xeon(R) Gold 5117 CPU @2.0GHz and NVIDIA RTX 3070 GPU. Each subgraph consists of three operators: padding (padding size is 1), C2D ($KH=KW=3, stride=1$), C2D ($KH=KW=1, stride=1$). The input height/width of subgraph\#1 is 7, while it is 14 for subgraph\#2. Besides, all the numbers of input/output channels are 512, except that the number of output channels of the latter C2D ($KH=KW=1$) in subgraph\#2 is 2048. We conduct two variants of \sys: \sys-FP and \sys-BP. \sys-FP will first tune C2D ($KH=KW=3$) and propagate its output layout to the input tensor of the latter C2D ($KH=KW=1$). While \sys-BP will first tune C2D ($KH=KW=1$) and propagate its input layout to the output tensor of the former C2D ($KH=KW=3$). Instead, \sys will tune the two C2Ds separately and insert a layout conversion operator between them according to the third constraint in \cref{subsec:layout_prop}. 

\begin{figure}
	\centering
	\includegraphics[width=0.46\textwidth]{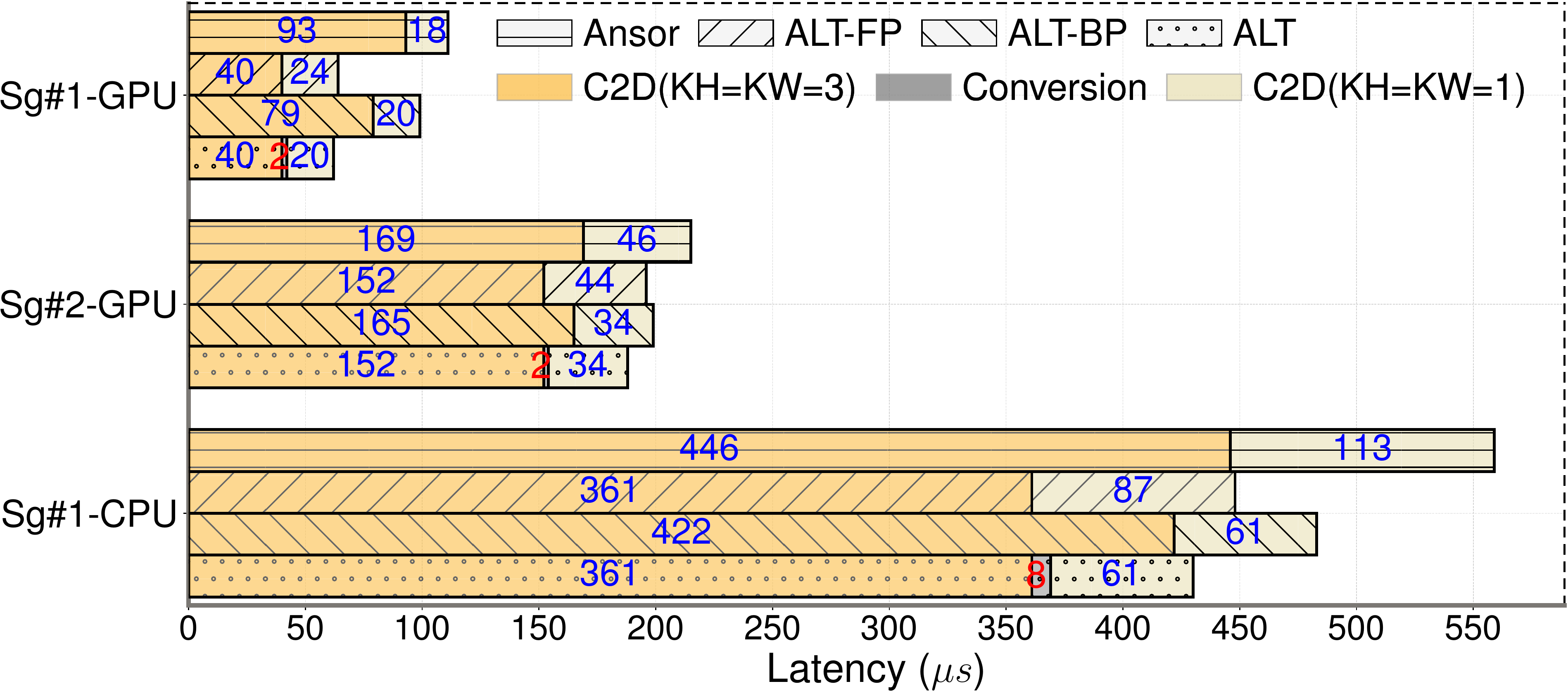} \hfill
	\caption{The overhead of layout propagation.}
	\vspace{-0.15in}
	\label{fig:prop_overhead}
\end{figure}

The profiling results are reported in \cref{fig:prop_overhead}, where we use Ansor as a reference point. We observe that \sys outperforms \sys-FP and \sys-WP. In other words, the best output layout of the C2D ($KH=KW=3$) is sub-optimal for the second C2D ($KH=KW=1$), and vice versa. Independent layout tuning for each complex operator brings more benefits while the layout conversion only incurs low overhead (2 microseconds for GPU and 8 microseconds for CPU). Combined with the results of \sys-WP in \cref{fig:exprim_network}, the fusion conflicts incur more overhead than layout conversions when performing layout transformation. We alleviate such two kinds of overheads by layout propagation and eschew the overhead of propagation itself by introducing necessary constraints.

\subsubsection{Parameter sensitivity:} 

\begin{figure}
	\centering
	\includegraphics[width=0.46\textwidth]{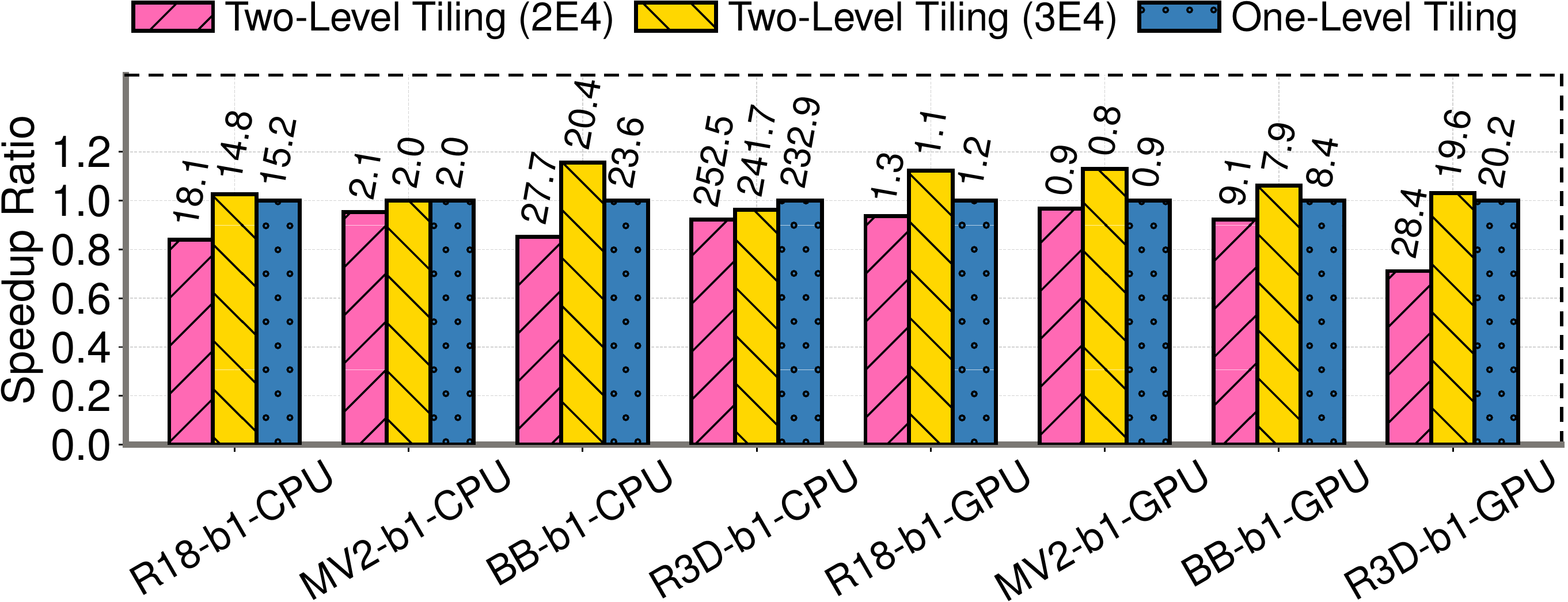} \hfill
	\caption{End-to-end performance of different settings.}
	\vspace{-0.15in}
	\label{fig:search_tradeoff_bench_v2}
\end{figure}

We study the parameter sensitivity by comparing the performance given different budget settings and search space sizes. We include three variants here: 1) two-level tiling templates with 20,000 budget;
2) two-level tiling templates but with 30,000 budget;
3) one-level layout tiling templates with 20,000 budget as the baseline (\textit{i.e.}, same as \cref{subsec:e2e_bench}).

The end-to-end performance in different settings is shown in \cref{fig:search_tradeoff_bench_v2}. The first variant expands the search space size while keeping the budget unchanged. Compared with it, the baseline illustrates 15\% performance improvement on average. By contrast, after setting the budget to 30,000, the second variant improves about 6\% performance over the baseline. Also, more improvements can be obtained if given a larger budget, since one-level tiling templates constitute a subset of the two-level variant. For the budget of 20,000 in \cref{subsec:e2e_bench}, one-level layout tiling templates yield a more effective trade-off between the final performance and the search space size. 
The budget of 20,000 to optimize a network typically costs 12-16 hours. But, it is affordable for practitioners as they only need to execute \sys once.
%
Additionally, these results demonstrate the scalability of the tuning space, which is hard to achieve in prior auto-tuning works.

\subsubsection{Case study:}
\label{subsubsec:case_study}

To understand how the joint tuning improves the loop performance, we perform loop optimization based on $NHWO$, $NOHW$, $N\frac{O}{o_t}HWo_t$, and $N\frac{H}{h_t} \frac{W}{w_t} \frac{O}{o_t} h_t w_t o_t$ on Intel CPU. We profiled a small computational graph, which contains several operators: padding (after padding, the tensor will have $N=1, I=3, H=W=230$), C2D ($O=64, KH=KW=7$, convolutional stride is 2), bias addition, and ReLU. This small graph is also the first layer of R18-b1. We set $o_t=16$ for $N\frac{O}{o_t}HWo_t$ ($i_t=3$ for the input tensor), while the searched layout has $h_t = 4, w_t = 16, o_t=16$ for $N\frac{H}{h_t} \frac{W}{w_t} \frac{O}{o_t} h_t w_t o_t$ ($i_t=1$ for the input tensor). {The platform is 48-core Intel(R) Xeon(R) Gold 5117 CPU @2.0GHz.} 

\begin{table}[!ht]
	\centering
	\small
	\caption{Profiling results based on several layouts.}
	\begin{tabular}{l@{\hskip 0.1in}c@{\hskip 0.05in}c@{\hskip 0.05in}c@{\hskip 0.05in}c@{\hskip 0.05in}c}
		\toprule
		\textbf{Layout (Conv \& Ker)} &\textbf{\#Inst.} &\textbf{\#L1-lds} &\textbf{\#L1-mis}
		&\textbf{\#L1-sts} & \textbf{Lat.}\\
		\midrule
		$NHWO$ \& $rsIO$& $509.4$ & $166.4$ & $9.7$ & $103.6$ & $0.34$\\
		\midrule
		$NOHW$ \& $OIrs$ & $626.9$ & $206.6$ & $4.5$ & $121.3$ & $0.49$\\
		\midrule
		$N\frac{O}{o_t}HWo_t$ \& $\frac{O}{o_t}\frac{I}{i_t}rsio$& $567.6$ & $193.6$ & $9.9$ & $112.9$ & $0.37$\\
		\midrule
		$N\frac{H}{h_t} \frac{W}{w_t} \frac{O}{o_t} h_t w_t o_t$ \& ... & $550.5$ & $174.3$ & $3.9$ & $106.2$ & $0.25$\\
		\bottomrule
	\end{tabular}
	\label{tab:prof_result}
\end{table}

The results are summarized in \cref{tab:prof_result}, where we abbreviate $(KH)(KW)$ to $rs$ for the weight tensor $Ker$. The latency (Lat.) is recorded in milliseconds and others are on a scale of $10^6$. We observe that for all layouts, except $NOHW$, their optimized loop nests prefer reusing input values by computing multiple output channels once with SIMD, thus reporting fewer instructions and fewer cache loads/stores than $NOHW$. Compared with $N\frac{O}{o_t}HWo_t$, $NHWO$ shows better data locality due to the larger tile size for the output channel. Specifically, $O=64$ in $NHWO$ yields a higher reuse rate than $o_t=16$ in $N\frac{O}{o_t}HWo_t$, as analyzed in \cref{subsec:space_build}. Further, $N\frac{H}{h_t} \frac{W}{w_t} \frac{O}{o_t} h_t w_t o_t$ achieves more efficient cache utilization  (only $2\%$ misses) than $NHWO$, due to the contiguous storage of intra-tile data elements after layout tiling.

\subsubsection{Other observations:} {Besides the profiled results, we observe that the $o_t$ parameter in the templates to tile output channels is often tuned as twice as the number of vector lanes that the platform supports when the spatial dimensions are not tiled.} Specifically, we observe that $o_t=32$ on Intel CPU, $o_t=8$ on NVIDIA GPU, and $o_t=8$ on ARM CPU frequently arise, although the number of vector lanes with float32 data types is 16 for AVX-512, 4 for CUDA, and 4 for NEON. This is different from many hand-tuned libraries. However, these results are not applicable to all configurations or platforms. By contrast, the methodology in our micro-benchmarks could help understand the optimized layout, and similar analysis can be conducted for other cases.
\presec\section{Related Work}
\vspace{0.05in}

\textbf{Deep learning compiler.} A variety of deep compilers have been developed. 
Halide \cite{ragan2013halide} and TVM \cite{chen2018tvm} decouple the operator description and schedule to simplify loop optimization. XLA \cite{leary2017xla}, Glow \cite{rotem2018glow}, nGraph \cite{cyphers2018intel}, and Relay \cite{roesch2018relay} develop graph-level representations to support layout selection, constant folding, etc. Rammer \cite{ma2020rammer} supports fine-grained operator fusion. CODE \cite{trainiti2021code} speeds up the ensemble of deep models. Cortex \cite{fegade2020cortex}, Nimble \cite{shen2020nimble}, DietCode \cite{zheng2022dietcode}, and CoRa \cite{fegade2022cora} focus on optimizing recursive/dynamic networks. TASO \cite{10.1145/3341301.3359630}, Tensat \cite{yang2021equality}, PET \cite{wang2021pet}, Unity \cite{unger2022unity}, and Ollie \cite{zheng2022ollie} perform subgraph substitutions to obtain a more efficient computational graph. Tensor Comprehension (TC) \cite{vasilache2018tensor}, Tiramisu \cite{baghdadi2019tiramisu}, MLIR \cite{lattner2021mlir}, and AKG \cite{zhao2021akg} integrate polyhedral techniques. Bolt \cite{xing2022bolt} provides support for tensor core by integrating CUTLASS \cite{cutlass}. SoyBean \cite{wang2018unifying} and Alpa \cite{zheng2022alpa} provide auto-tuning support for inter- and intra-operator parallelism in distributed scenarios. UNIT\cite{weng2021unit}, AMOS \cite{zheng2022amos}, and TensorIR \cite{feng2022tensorir} provide support for tensorization on tensor accelerators. SparTA \cite{zheng2022sparta} and SparseTIR \cite{ye2022sparsetir} introduce representation for sparse tensors. Compared with \sys, the layout auto-tuning, together with the joint data layout and loop optimization, is limited in these works. For instance, TC and Tiramisu require developers to transform data buffers manually. Although Relay and TVM can insert layout conversion operators between C2Ds with different predefined layouts (\textit{e.g.}, $NOHW$, $NHWO$, etc.), each layout combination requires a manual re-implementation of operators. By contrast, \sys supports generic graph-level layout auto-tuning with feedback from operator-level optimization.


\textbf{Layout and loop tuning.} Many systems try to improve the performance with layout transformation \cite{zhang2015optimizing, li2016optimizing, chen2018learning, liu2019optimizing, liu2020cocopie, zheng2020ansor, chou2020automatic, phothilimthana2021flexible, edo2021boveda}. For instance, \cite{zhang2015optimizing, edo2021boveda} optimize data layouts for FPGA design. \cite{li2016optimizing, phothilimthana2021flexible} suggests to choose layouts among $NHWO$, $NOHW$, etc. \cite{liu2020cocopie, chou2020automatic} tightly couples it with the sparse computation. Compared with \sys, they lack versatility and are limited to a few tuning options.
By contrast, the systems in \cite{chen2018learning, zheng2020ansor} can typically set the $o_t$ parameter in $N\frac{O}{o_t}HWo_t$ layout after integrating NeoCPU \cite{liu2019optimizing}. However, they have limitations: 1) switching to another kind of layout (\textit{e.g.}, a different reorder, or the overlapped tiling in \sys) still requires manually rewriting operators and even the templates of loop tuning, due to the coupling among data storage, operator implementation, and the loop-tuning templates; 2) $o_t$ is typically predetermined, while in Ansor \cite{zheng2020ansor} is set via resolving the loop tiling configurations after loop tuning, as a packing technique and only for constant tensors, hence no joint tuning. 
\sys addresses the two limitations via 1) the generic layout transformation submodule, which requires no re-implementation, and is also independent of the loop transformation to achieve the decoupling; 2) an auto-tuning module at a higher level to orchestrate the cross-layer joint tuning while guaranteeing efficiency. 
As for recent loop optimization techniques \cite{zheng2020flextensor, ahn2020chameleon, li2021analytical, wang2021tuna, steiner2021, baghdadi2021deep, ding2021ios, zhao2022apollo, zheng2022astitch, yu2021lorien, steiner2021value, ahrens2021asymptotic, zheng2021tenset, zhu2022roller}, such as delicate cost models \cite{li2021analytical, wang2021tuna, baghdadi2021deep, ahrens2021asymptotic}, aggressive operator fusion \cite{niu2021dnnfusion, ma2020rammer, ding2021ios, zhao2022apollo, zheng2022astitch, li2022automatic}, and micro-kernel construction \cite{zhu2022roller}, they are complementary to \sys.

\section{Conclusion}
In this paper, we propose \sys, a compiler that jointly performs graph-level data layout optimization and operator-level loop optimization for deep models.
\sys provides a generic transformation module for low-cost layout and loop manipulation. It further integrates an auto-tuning module for bidirectional and unified layout and loop tuning.
Experiments show that \sys outperforms state-of-the-art vendor libraries and auto-tuning frameworks.
\bibliography{main}
\bibliographystyle{plain}

\end{document}